\documentclass{article} 

\usepackage[dvipsnames]{xcolor}

\usepackage{colm2024_conference}

\colmfinalcopy 

\usepackage{microtype}
\usepackage{hyperref}
\usepackage{url}
\usepackage{booktabs}
\usepackage{enumitem}
\usepackage[accsupp]{axessibility} 
\usepackage{wrapfig}

\usepackage{amsmath,amsfonts,bm}

\def\eqref#1{equation~\ref{#1}}

\def\1{\bm{1}}

\DeclareMathAlphabet{\mathsfit}{\encodingdefault}{\sfdefault}{m}{sl}
\SetMathAlphabet{\mathsfit}{bold}{\encodingdefault}{\sfdefault}{bx}{n}

\usepackage{microtype}
\usepackage{graphicx}
\usepackage{booktabs}
\usepackage{boldline}
\usepackage{multirow}
\usepackage{color}
\usepackage{colortbl}
\usepackage{wrapfig}

\definecolor{lightblue}{rgb}{0.90, 0.95, 0.98}
\newcolumntype{a}{>{\columncolor{lightblue}}c}

\newcommand{\method}{\textsc{VideoDirectorGPT}}
\newcommand{\videomodule}{Layout2Vid}
\newcommand{\videoplan}{\textit{video plan}}
\newcommand{\corefdata}{Coref-SV}
\newcommand{\modelscope}{ModelScopeT2V}

\usepackage[capitalize]{cleveref}
\crefname{section}{Sec.}{Secs.}
\Crefname{section}{Section}{Sections}
\Crefname{table}{Table}{Tables}
\crefname{table}{Tab.}{Tabs.}

\usepackage[page,header]{appendix}
\usepackage{titletoc}
\usepackage{tocloft}
\usepackage{xspace}

\makeatletter
\DeclareRobustCommand\onedot{\futurelet\@let@token\@onedot}
\def\@onedot{\ifx\@let@token.\else.\null\fi\xspace}

\def\eg{\emph{e.g}\onedot} 
\def\ie{\emph{i.e}\onedot} 
 
\def\etc{\emph{etc}\onedot}

\makeatother

\definecolor{Red}{rgb}{0.6,0,0}
\definecolor{Blue}{rgb}{0,0,0.8}
\definecolor{Green}{rgb}{0.2,0.8,0}
\definecolor{airforceblue}{rgb}{0.36, 0.54, 0.66}
\definecolor{ao(english)}{rgb}{0.0, 0.5, 0.0}
\definecolor{azure(colorwheel)}{rgb}{0.0, 0.5, 1.0}
\definecolor{crimson}{rgb}{0.86, 0.08, 0.24}
\definecolor{darkcerulean}{rgb}{0.03, 0.27, 0.49}
\definecolor{cobalt}{rgb}{0.0, 0.28, 0.67}
\definecolor{rosegold}{rgb}{0.72, 0.43, 0.47}
\definecolor{orange-red}{rgb}{1.0, 0.27, 0.0}
\definecolor{mountainmeadow}{rgb}{0.19, 0.73, 0.56}
\definecolor{malachite}{rgb}{0.04, 0.85, 0.32}
\definecolor{darkblue}{rgb}{0.0, 0.0, 0.55}
\definecolor{customblue}{rgb}{0.2, 0.35, 0.8}

\hypersetup{colorlinks,linkcolor={blue},citecolor={mountainmeadow},urlcolor={magenta}}

\title{\method{}: Consistent Multi-Scene Video\\ Generation via LLM-Guided Planning}

\author{Han Lin
\quad
Abhay Zala
\quad
Jaemin Cho
\quad
Mohit Bansal\\
UNC Chapel Hill\\
\texttt{\{hanlincs, aszala, jmincho, mbansal\}@cs.unc.edu} \\
{\tt \normalsize \href{https://videodirectorgpt.github.io/}{https://videodirectorgpt.github.io/}}
}

\begin{document}

\maketitle

\begin{abstract}
Recent text-to-video (T2V) generation methods have seen significant advancements. However, the majority of these works focus on producing short video clips of a single event (i.e., single-scene videos).
Meanwhile, recent large language models (LLMs) have demonstrated their capability in generating layouts and programs to control downstream visual modules.
This prompts an important question: {\sl can we leverage the knowledge embedded in these LLMs for temporally consistent long video generation?} 
In this paper, we propose \method{}, a novel framework for consistent multi-scene video generation that uses the knowledge of LLMs for video content planning and grounded video generation.
Specifically, given a single text prompt, we first ask our video planner LLM (GPT-4) to expand it into a `\videoplan{}', which includes the scene descriptions, the entities with their respective layouts, the background for each scene, and consistency groupings of the entities. 
Next, guided by this \videoplan{}, our video generator, named \videomodule{}, has explicit control over spatial layouts and can maintain temporal consistency of entities across multiple scenes, while being trained only with image-level annotations.
Our experiments demonstrate that our proposed \method{} framework
substantially improves layout and movement control in both single- and multi-scene video generation and can generate multi-scene videos with consistency, while achieving competitive performance with SOTAs in open-domain single-scene T2V generation.
Detailed ablation studies, including dynamic adjustment of layout control strength with an LLM and video generation with user-provided images, confirm the effectiveness of each component of our framework and its future potential.
\end{abstract}

\section{Introduction}
\label{sec:intro}

Text-to-video (T2V) generation has achieved rapid progress following the success of text-to-image (T2I) generation.
Most works in T2V generation focus on producing short videos (e.g., 16 frames at 2fps)
from the given text prompts~\citep{wang2023modelscope, he2022latent, ho2022imagen, singer2022make, zhou2022magicvideo}.
Recent studies on long video generation \citep{blattmann2023align, yin2023nuwa, villegas2022phenaki, he2023animate} aim at generating long videos of a few minutes with holistic visual consistency.
Although these works could generate longer videos, the generated videos often display 
the continuation or repetitive patterns of a single action (e.g., driving a car) instead of transitions and dynamics of multiple changing actions/events (e.g., five steps about how to make caraway cakes).
Meanwhile, large language models (LLMs)~\citep{brown2020language, openai2023gpt4, touvron2023llama1, touvron2023llama, chowdhery2022palm} have demonstrated their capability in generating layouts and programs to control visual modules~\citep{surismenon2023vipergpt,gupta2023visual}, especially image generation models~\citep{cho2023visual,feng2023layoutgpt}.
This raises an interesting question:
{\sl Can we leverage the knowledge embedded in these LLMs for planning consistent multi-scene video generation?}

\begin{figure}[t]
\begin{center}
    \includegraphics[width=.99\linewidth]{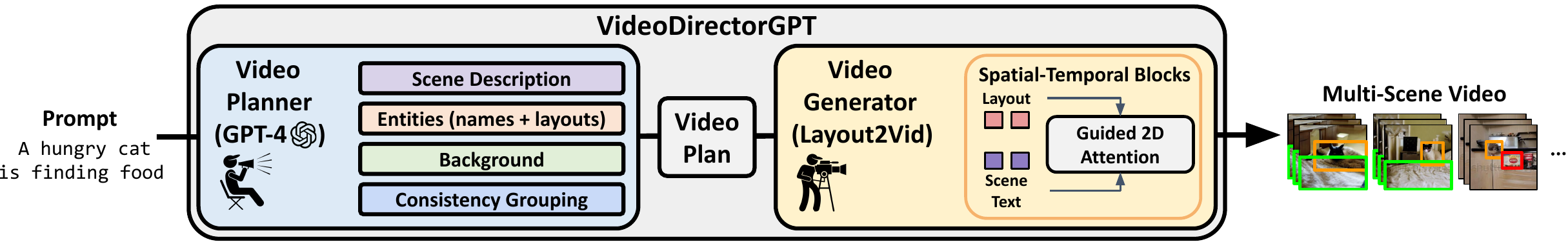}
    \vspace{-2mm}
    \caption{
    \textbf{
    \method{} framework.}
    First,
    we employ \textbf{GPT-4 as a video planner} to craft a multi-component \textbf{\videoplan{}}.
    Next,
    we utilize \textbf{\videomodule{}}, a grounded video generation module, to render multi-scene videos with layout and consistency control. 
    \vspace{-20pt}
    }
\label{fig:teaser}
\end{center}
\end{figure}

In this work, we introduce \textbf{\method{}}, a novel framework for consistent multi-scene video generation. As illustrated in \cref{fig:teaser},
\method{} decomposes the T2V generation task into two stages:
\textbf{video planning} and \textbf{video generation}.
For the first video planning stage (see \cref{fig:teaser} \textcolor{ProcessBlue}{blue} part), we employ an LLM to generate a \videoplan{}, which is an overall plot of the video with multiple scenes, each consisting of a text description of the scene and entity names/layouts, and a background. It also consists of consistency groupings 
of specific entities that re-appear across scenes.
For the second video generation stage (see \cref{fig:teaser} \textcolor{Dandelion}{yellow} part), we introduce \videomodule{}, a novel grounded video generation module that generates multi-scene videos from the \videoplan{}.
Our framework provides the following strengths:
(1) employing an LLM to write a \videoplan{} that guides the generation of videos {with multiple scenes from a single text prompt},
(2) layout control in video generation by only using image-level layout annotations, and (3) generation of visually consistent entities across multiple scenes.

\begin{figure}[t]
\begin{center}
    \includegraphics[width=0.99\linewidth]{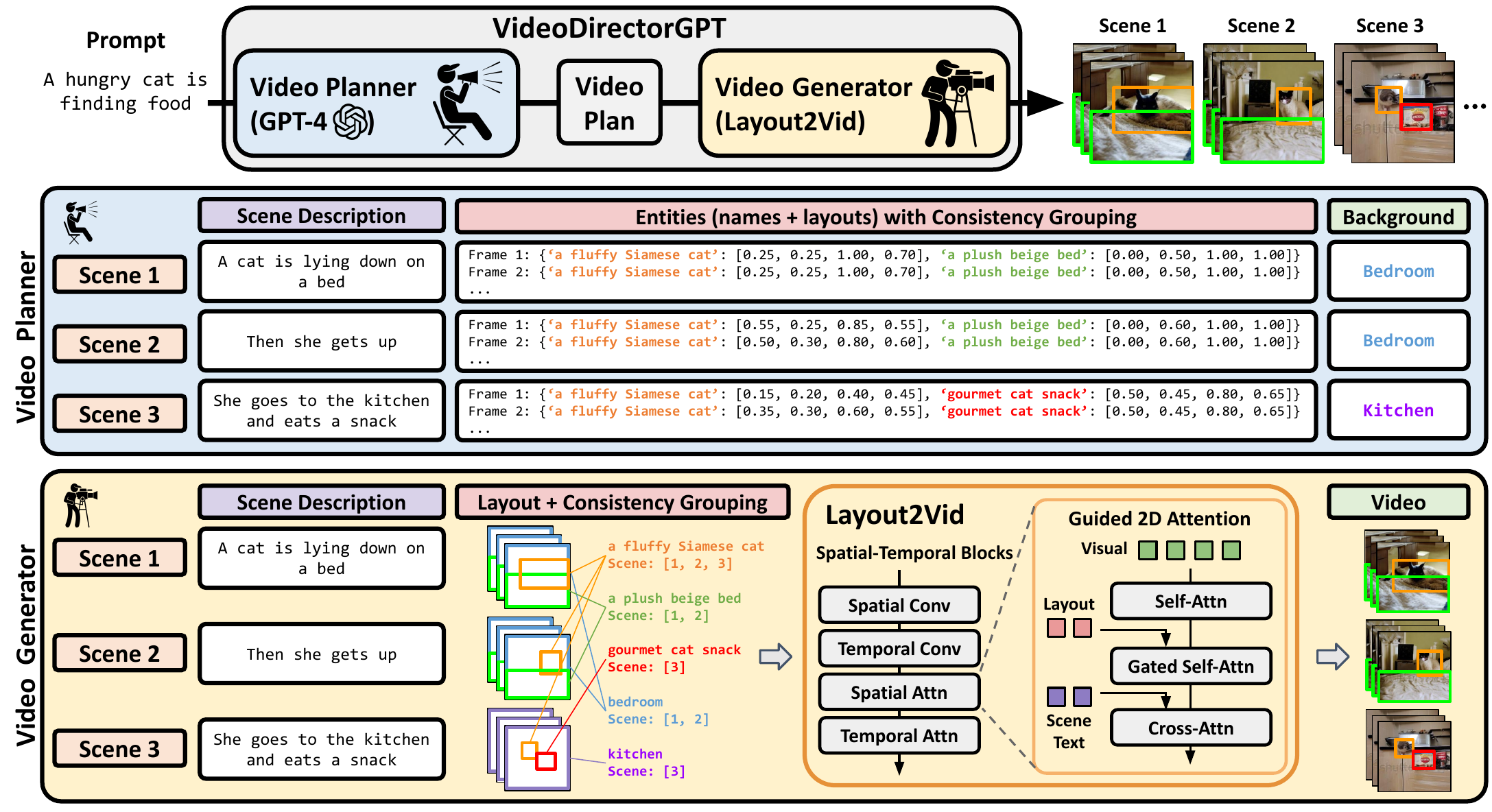}
    \vspace{-2mm}
    \caption{
    \textbf{
    \method{} details.
    }
    \textbf{1st stage}:
    we employ the \textbf{LLM as a video planner} to craft a \textbf{\videoplan{}},
    which provides an overarching plot for multi-scene videos.
    \textbf{2nd stage}:
    we utilize \textbf{\videomodule{}}, a grounded video generation module, to render videos based on the \videoplan{}.
    (\cref{sec:grounded_video_generation}).
    \vspace{-10pt}
    }
\label{fig:framework}
\end{center}
\end{figure}

In the first stage, video planning (\cref{sec:llm_guiding}),
we employ an LLM (e.g., GPT-4~\citep{openai2023gpt4}) as a video planner to generate a \textbf{\videoplan{}},
a multi-component video script with multiple scenes to guide the downstream video synthesis process.
Our \videoplan{} consists of four components: 
(1) multi-scene descriptions,
(2) entities (names and their 2D bounding boxes),
(3) background,
and
(4) consistency groupings (scene indices for each entity indicating where they should remain visually consistent).
We generate the \videoplan{} in two steps by prompting an LLM with different in-context examples. 
In the first step, we expand a single text prompt into multi-step scene descriptions with an LLM, 
where each scene is described with a text description, a list of entities, and a background (see \cref{fig:framework} \textcolor{ProcessBlue}{blue} part for details).
We also prompt the LLM to generate additional information for each entity (e.g., color, attire, etc.), and group together entities across frames and scenes, which will help guide consistency during the video generation stage.
In the second step, we expand the detailed layouts of each scene with an LLM by generating the bounding boxes of the entities per frame, given the list of entities and scene description.
This overall `\videoplan{}' guides the downstream video generation.

In the second stage, video generation (\cref{sec:grounded_video_generation}),
we introduce \videomodule{},
a grounded video generation module to render videos based on the generated \videoplan{} (see \textcolor{Dandelion}{yellow} part of \cref{fig:framework}).
For the grounded video generation module, we build upon \modelscope{}~\citep{wang2023modelscope}, an off-the-shelf T2V generation model, by freezing its original parameters and adding spatial/consistency control of entities through a small set of trainable parameters (13\% of total parameters) through the gated-attention module~\citep{li2023gligen}. 
This enables our \videomodule{} to be trained solely on layout-annotated images, thus bypassing the need for expensive training on annotated video datasets.
To preserve the identity of entities across scenes,
we use shared representations for the entities within the same consistency group.
We also propose to use a joint image+text embedding as entity grounding conditions 
which we find more effective than the existing text-only approaches~\citep{li2023gligen} in entity identity preservation (see \cref{sec:ablation_study}).
Overall, our \videomodule{} avoids expensive video-level training, improves the object layout and movement control, and preserves objects cross-scene temporal consistency.

We conduct experiments on both single-scene and multi-scene video generation. 
Experiments show that our \method{} demonstrates better layout skills (object, count, spatial, scale) 
and object movement control compared with \modelscope{}~\citep{wang2023modelscope} as well as more recent video generation models including  AnimateDiff~\citep{guo2023animatediff}, I2VGen-XL~\citep{zhang2023i2vgen}, and SVD~\citep{blattmann2023stable}. In addition, \method{} is capable of generating multi-scene videos with visual consistency across scenes, and competitive with SOTAs on single-scene open-domain T2V generation (\cref{sec:single_scene_video_gen} and \cref{sec:multi_scene_video_gen}).
Detailed ablation studies, including dynamic adjustment of layout control strength, video plans generated from different LLMs, and video generation with user-provided images confirm the effectiveness and capacity of our framework (\cref{sec:additional_analysis} and \cref{sec:ablation_study}).

Our main contributions can be summarized as follows:
\begin{itemize}[leftmargin=1em]
    \setlength{\itemsep}{0pt}
    \setlength{\parskip}{0pt}
    \setlength{\parsep}{0pt}
    \item A new T2V generation framework \method{} with two stages: video content planning and grounded video generation, which is capable of generating a multi-scene video from a single text prompt.
    \item We employ LLMs to generate a multi-component `\videoplan{}'
    with detailed scene descriptions, entity layouts, and entity consistency groupings 
    to guide video generation.
    \item We introduce \videomodule{}, a novel grounded video generation module, which brings together image/text-based layout control ability and entity-level temporal consistency. Our \videomodule{} can be trained efficiently using only image-level layout annotations.
    \item Empirical results demonstrate that our framework can accurately control object layouts and movements, and generate temporally consistent multi-scene videos.
\end{itemize}

\section{Related Works}
\label{sec:related_works}

\noindent\textbf{Text-to-video generation.}
Training a T2V generation model from scratch is computationally expensive.
Recent work often leverages pre-trained text-to-image (T2I) generation models such as Stable Diffusion~\citep{rombach2022high} by fine-tuning them on text-video pairs~\citep{wang2023modelscope, blattmann2023align}.
While this warm-start strategy enables high-resolution video generation, it comes with the limitation of only generating short video clips, as T2I models lack the ability to maintain consistency through long videos. 
On the other hand, recent works on long video generation \citep{blattmann2023align, yin2023nuwa, villegas2022phenaki, he2023animate} aim at generating long videos of several minutes.
However, the generated videos often display the continuation or repetitive actions 
instead of transitions of multiple actions/events. 
In contrast, our \videomodule{} infuses layout control and multi-scene temporal consistency into a pretrained T2V generation model via data and parameter-efficient training while preserving its original visual quality.

\vspace{5pt}
\noindent\textbf{Bridging text-to-image generation with layouts.}
To achieve interpretable and controllable generation, a line of research decomposes the T2I generation task into two stages: text-to-layout generation, and layout-to-image generation.
While early models train the layout generation module from scratch~\citep{hong2018inferring,tan2019text2scene,li2019object,Liang2022LayoutBridgingTS},
recent methods employ pretrained LLMs to leverage their knowledge in generating image layouts from text~\citep{cho2023visual,feng2023layoutgpt,qu2023layoutllmt2i}.
To the best of our knowledge, our work is the first to utilize LLMs to generate structured video plans 
from text, enabling accurate and controllable long video generation.

\begin{figure}[t]
\begin{center}
    \includegraphics[width=.99\linewidth]{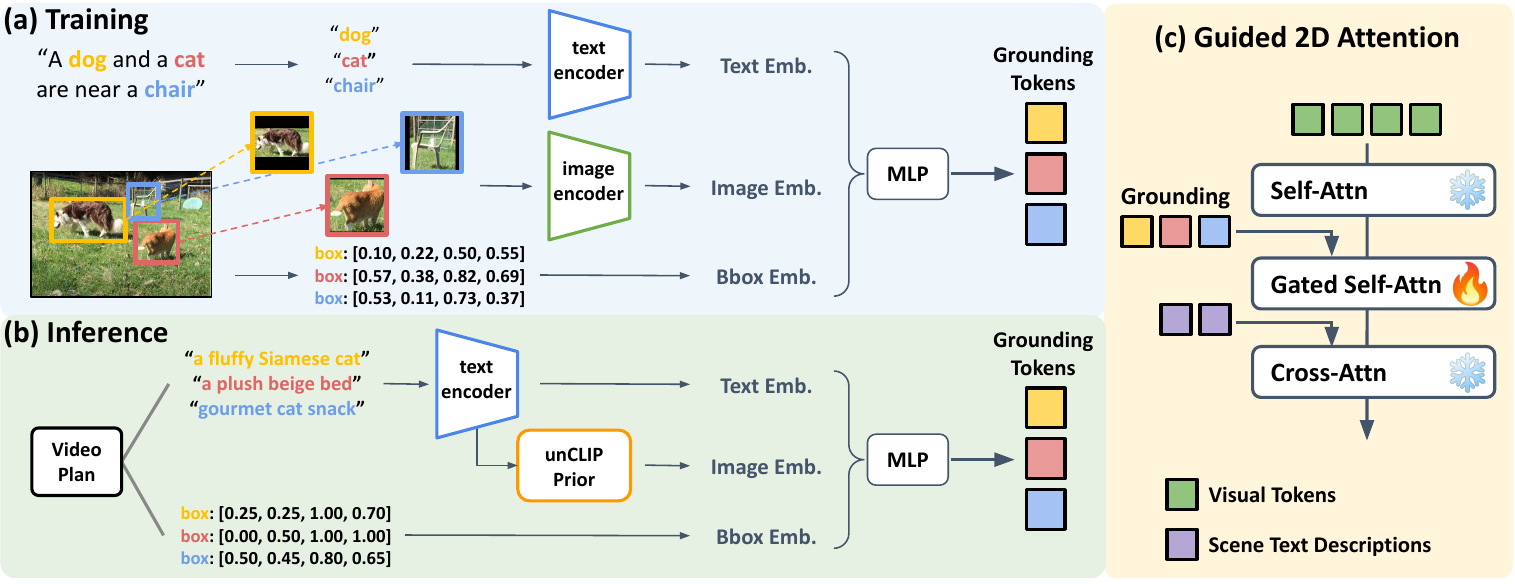}
    \vspace{-3mm}
    \caption{
    \textbf{Illustration of \videomodule{}.}
    \textbf{(a) Training.}
    Our \videomodule{}
    is efficiently trained on image-level bounding box annotations.
    We construct entity grounding tokens 
    with an MLP that takes text, image, and bounding box
    embeddings as inputs.
    \textbf{(b) Inference.}
    we obtain image embeddings of entities from their text descriptions with unCLIP Prior.
    To ensure temporal consistency, we use the same entity embeddings across scenes.
    \textbf{(c) Guided 2D Attention.}
    We modulate the \textcolor{black}{visual representation} with \textcolor{black}{grounding tokens} and \textcolor{black}{text tokens}.
    }
\label{fig:img_embedding}
\vspace{-4mm}
\end{center}
\end{figure}

\section{\method{}}
\label{sec:method}

\subsection{Video Planning: Video Plans with LLMs}
\label{sec:llm_guiding}

\vspace{5pt}
\noindent\textbf{Video plan.}
As illustrated in the \textcolor{ProcessBlue}{blue} part of \cref{fig:framework}, GPT-4~\citep{openai2023gpt4} 
acts as a planner, providing a detailed \textit{\videoplan{}} 
to guide the video generation.
Our \videoplan{} has four components:
(1) \textbf{multi-scene descriptions}: a sentence describing each scene, 
(2) \textbf{entities}: names and their 2D bounding boxes, 
(3) \textbf{background}: text description of the location of each scene,
and (4) \textbf{consistency groupings}:
scene indices for each entity indicating where they should remain visually consistent.
The \videoplan{} is generated in two steps by prompting GPT-4 independently. 
The input prompt template for each step is displayed in \cref{appendix:prompt_templates}.

\vspace{5pt}
\noindent\textbf{Step 1: Generating multi-scene descriptions, entity names, and consistency groupings.}
In the first step, we employ GPT-4 to expand a single text prompt into a multi-scene \videoplan{}.
Next, we group entities and backgrounds that appear across different scenes using an exact match. For instance, if the `chef' appears in scenes 1-4 and `oven' only appears in scene 1, we form the entity consistency groupings as \texttt{\{chef:[1,2,3,4], oven:[1]\}}. 
In the subsequent video generation stage, we use the
shared representations for the same entity consistency groups to maintain their temporally consistent appearances (see \cref{sec:grounded_video_generation}).

\vspace{5pt}
\noindent\textbf{Step 2: Generating entity layouts for each scene.}
In the second step,
we generate a list of bounding boxes for the entities in each frame based on the entities and the scene description. For each scene, we produce layouts for 9 frames, then linearly interpolate to gather information for denser frames (\eg{}, 16 frames per scene).
In line with VPGen~\citep{cho2023visual}, we adopt the $[x_0, y_0, x_1, y_1]$ format for bounding boxes, where each coordinate is normalized to fall within the range [0,1].
For in-context examples, we present 0.05 as the minimum unit for the bounding box, equivalent to a 20-bin quantization.

\subsection{Video Generation: Generating Videos from Video Plans with \videomodule{}}
\label{sec:grounded_video_generation}

\vspace{5pt}
\noindent\textbf{Preliminaries.}
Given a $T$ frame video $\bm{x}\in\mathbb{R}^{T\times 3 \times H \times W}$ with video caption $\bm{c}$, we first uses an autoencoder to encode the video into latents $\bm{z}=\mathcal{E}(\bm{x})\in\mathbb{R}^{T\times C \times H' \times W'}$, where $C=4$ represents the number of latent channels, and $H'=H/8$, $W'=W/8$ represent the spatial dimension of the latents. 
In Latent diffusion model (LDM) \citep{rombach2022high}, the forward process is a fixed diffusion process which gradually adds noise to the latent variable $\bm{z}$:
$q(\bm{z}_t|\bm{z}_{t-1})=N(\bm{z}_t; \sqrt{1-\beta_t}\bm{z}_{t-1}, \beta_t \bm{I})$,
where $\beta_t\in(0,1)$ is the variance schedule with $t\in\{1, ..., T\}$.
The reverse process gradually produces less noisy samples $\bm{z}_{T-1}, \bm{z}_{T-2}, ..., \bm{z}_0$ starting from $\bm{z}_{T}$ through a learnable denoiser model $\bm{\epsilon_\theta}$. With t as a timestep uniformly sampled from $\{1, ..., T\}$, the training objective of our \videomodule{} can be expressed as: 
$\min_{\bm{\theta}} L_{\text{LDM}}=E_{\bm{z}, \bm{\epsilon}\sim N(0, \bm{I}), t}\|\bm{\epsilon} - \bm{\epsilon_{\theta}}(\bm{z}_t, t, \bm{c})\|_2^2$.

\vspace{5pt}
\noindent\textbf{\videomodule{}: Layout-guided T2V generation.}
\label{sec:network_structure}
We implement \videomodule{} by integrating layout control capability into \modelscope{}~\citep{wang2023modelscope}, a public T2V generation model based on Stable Diffusion~\citep{rombach2022high}.
The diffusion UNet in \modelscope{} consists of a series of spatio-temporal blocks, each containing four modules: spatial convolution, temporal convolution, spatial attention, and temporal attention.
Compared with \modelscope{}, our \videomodule{}
enables layout-guided video generation 
with explicit spatial control over a list of entities represented by their bounding boxes, as well as visual and text content. 
We build upon the 2D attention 
to create the guided 2D attention (see \cref{fig:network} in \cref{appendix:layout2vid}).
As shown in \cref{fig:img_embedding} (c), the guided 2D attention takes two conditional inputs to modulate the visual latent representation:
(1) \textcolor{black}{grounding tokens}, conditioned with gated self-attention~\citep{li2023gligen},
and
(2) \textcolor{black}{text tokens} that describe the current scene, conditioned with cross-attention.

\vspace{5pt}
\noindent\textbf{Temporally consistent entity grounding with image+text embeddings.}
While previous layout-guided T2I generation models commonly used only the CLIP text embedding for layout control~\citep{li2023gligen,yang2022reco},
we use the CLIP image embedding in addition to the CLIP text embedding for entity grounding. We demonstrate in our ablation studies (see \cref{sec:ablation_study}) that this method is more effective than using either text-only or image-only grounding.
As depicted below, the grounding token for the $i^{th}$ entity, $\bm{h}_i$, is a 2-layer MLP which fuses
CLIP image embeddings $\bm{f}_{\text{img}}(e_i)$,
CLIP text embeddings $\bm{f}_{\text{text}}(e_i)$,
and Fourier features~\citep{mildenhall2021nerf} of the bounding box $\bm{l}_i = [x_0, y_0, x_1, y_1]$.
We use learnable linear projection layers, denoted as $\bm{P}_{\text{img/text}}$, on the visual/text features, which we found helpful for faster convergence during training.
\label{eq:our_grounding_token}
$$
\bm{h}_i = \text{MLP}(
\bm{P}_{\text{img}}(\bm{f}_{\text{img}}(e_i)),
\bm{P}_{\text{text}}(\bm{f}_{\text{text}}(e_i)),
\text{Fourier}(\bm{l}_i))
$$
The training and inference procedure of our \videomodule{} are presented in \cref{fig:img_embedding} parts (a) and (b) respectively.
During training, the image embeddings $\bm{f}_{\text{img}}(e)$ are obtained by encoding the image crop of the entities with CLIP image encoder.
During inference, since all the inputs are in text format (e.g., from the \videoplan{}), we employ Karlo~\citep{kakaobrain2022karlo-v1-alpha}, a public implementation of unCLIP Prior~\citep{ramesh2022hierarchical} to transform the CLIP text embedding into its corresponding CLIP image embedding.
Moreover, 
our image embeddings can also be obtained from user-provided exemplars during inference by simply encode the images with the CLIP image encoder.

\vspace{5pt}
\noindent\textbf{Parameter and data-efficient training.}
During training, we only update the parameters of the guided 2D attention (13\% of total parameters) to inject layout guidance capabilities into the pretrained T2V backbone while preserving its original video generation capabilities.
Such strategy allows us to efficiently train the model with only image-level layout annotations,
while still equipped with multi-scene temporal consistency via shared entity grounding tokens. Training and inference details are shown in \cref{appendix:layout2vid}.

\section{Experimental Setup}\label{sec:experiments}

\noindent\textbf{Evaluated models.} 
We primarily compare our \method{} with \modelscope{}, as our \videomodule{} uses its frozen weights and only trains a small set of new parameters to add spatial and consistency control. 
Additionally, we present comparisons with other T2V generation models (see \cref{sec:experiments_appendix} for baseline model details) on MSR-VTT and UCF-101 datasets~\citep{xu2016msr-vtt, soomro2012ucf101}.

\vspace{5pt}
\noindent\textbf{Prompts for single-scene video generation.}
For single-scene videos, 
we (1) evaluate layout control via VPEval Skill-based prompts~\citep{cho2023visual}, 
(2) assess object dynamics through ActionBench-Direction prompts adapted from ActionBench-SSV2~\citep{wang2023paxion}, 
and (3) examine open-domain video generation using MSR-VTT and UCF-101~\citep{xu2016msr-vtt, soomro2012ucf101}.
We introduce ActionBench-Direction prompts by sampling video captions from ActionBench-SSV2~\citep{wang2023paxion} and balancing the distribution of movement directions. 
Prompt preparation details are presented in \cref{sec:experiments_appendix}.

\vspace{5pt}
\noindent\textbf{Prompts for multi-scene video generation.} 
For multi-scene video generation,
we experiment with two types of input prompts:
(1) a list of sentences describing events -- ActivityNet Captions~\citep{krishna2017dense} and
\corefdata{} prompts based on Pororo-SV~\citep{li2018storygan}, and
(2) a single sentence from which models generate multi-scene videos -- HiREST~\citep{zala2023hierarchical}.
\corefdata{} is a new multi-scene text description dataset
that we propose to 
evaluate the visual consistency of objects across multi-scene videos.
Prompt preparation details are presented in \cref{sec:experiments_appendix}.

\vspace{5pt}
\noindent\textbf{Automated evaluation metrics.} Following previous works \citep{hong2022cogvideo, wu2022nuwa1, wang2023modelscope},
we use {FID}~\citep{heusel2017gans}, {FVD}~\citep{unterthiner2019fvd}, and {IS}~\citep{salimans2016improved} scores as video quality metrics, and {CLIPSIM}~\citep{wu2021godiva} score
for text-video alignment.
Given that CLIP fails to faithfully capture detailed semantics such as spatial relations, object counts, and actions
in videos~\citep{otani2023verifiable,Cho2023DallEval,cho2023visual,hu2023tifa},
we further propose the use of the following fine-grained metrics:
\begin{itemize}[leftmargin=1em]
    \item \textbf{VPEval accuracy}: we employ it for the evaluation of VPEval Skill-based prompts (object, count, spatial, scale), which is based on running skill-specific evaluation programs that execute relevant visual modules~\citep{cho2023visual}.
    Since the VPEval accuracy described above does not cover temporal information, we propose a metric that takes into account temporal information as well as spatial layouts for ActionBench-Direction prompts.
    \item \textbf{Object movement direction accuracy}: we introduce this new metric to evaluate ActionBench-Direction prompts, which takes both temporal information and spatial layouts into consideration. 
    To accomplish this, we assess whether the target objects in the generated videos move in the direction described in the prompts.
The start/end locations of objects are obtained by detecting objects with GroundingDINO~\citep{liu2023grounding} on the first/last video frames.
We then evaluate whether the $x$ (for movements left or right) or $y$ (for movements up or down) coordinates of the objects have changed correctly and assign a binary score of 0 or 1 based on this evaluation.
For instance, given the prompt ``pushing a glass from left to right'' and a generated video, we identify a `glass' in both the first and last video frames.
We assign a score of 1 if the $x$-coordinate of the object increases by the last frame; otherwise, we assign a score of 0.
    \item \textbf{Multi-scene object temporal consistency}: we propose this new metric for the evaluation of ActivityNet Captions and \corefdata{}, which measures the consistency of the visual appearance of a target object across different scenes.
    We also introduce a new metric to measure the consistency of the visual appearance of a target object across different scenes. 
Specifically, we first detect the target object using GroundingDINO from the center frame of each scene video.
Then, we extract the CLIP (ViT-B/32) image embedding from the crop of the detected bounding box.
We calculate the multi-scene object consistency metric by averaging the CLIP image embedding similarities across all adjacent scene pairs:
$\frac{1}{N} \sum_{n=1}^{N-1} cos(\text{CLIP}^{\text{img}}_{n}, \text{CLIP}^{\text{img}}_{n+1})$,
where $N$ is the number of scenes,
$cos(\cdot,\cdot)$ is cosine similarity,
and $\text{CLIP}^{\text{img}}_{n}$ is the CLIP image embedding of the target object in $n$-th scene.

\end{itemize}

\vspace{3pt}
\noindent\textbf{Human evaluation.}
We conduct a human evaluation on the multi-scene videos generated by both our \method{} and \modelscope{} on the \corefdata{} dataset.
Since we know the target entity and its co-reference pronouns in the \corefdata{} prompts, we can compare the temporal consistency of the target entities across scenes.
We evaluate the human preference between videos from two models in each category of Quality, Text-Video Alignment, and Object Consistency.
See \cref{sec:experiments_appendix} for setup details. 

\noindent\textbf{Step-by-step error analysis.}
We conduct an error analysis at each step of our single-sentence to multi-scene video generation pipeline on the HiREST dataset.
We analyze the generated multi-scene text descriptions, layouts, and entity consistency groupings to evaluate our video planning stage, and examine the final video to evaluate the video generation stage.

\section{Results and Analysis}
\label{sec:results}

\subsection{Single-Scene Video Generation}\label{sec:single_scene_video_gen}

\begin{figure}[h]
\vspace{-5pt}
  \centering
\includegraphics[width=.9\linewidth]{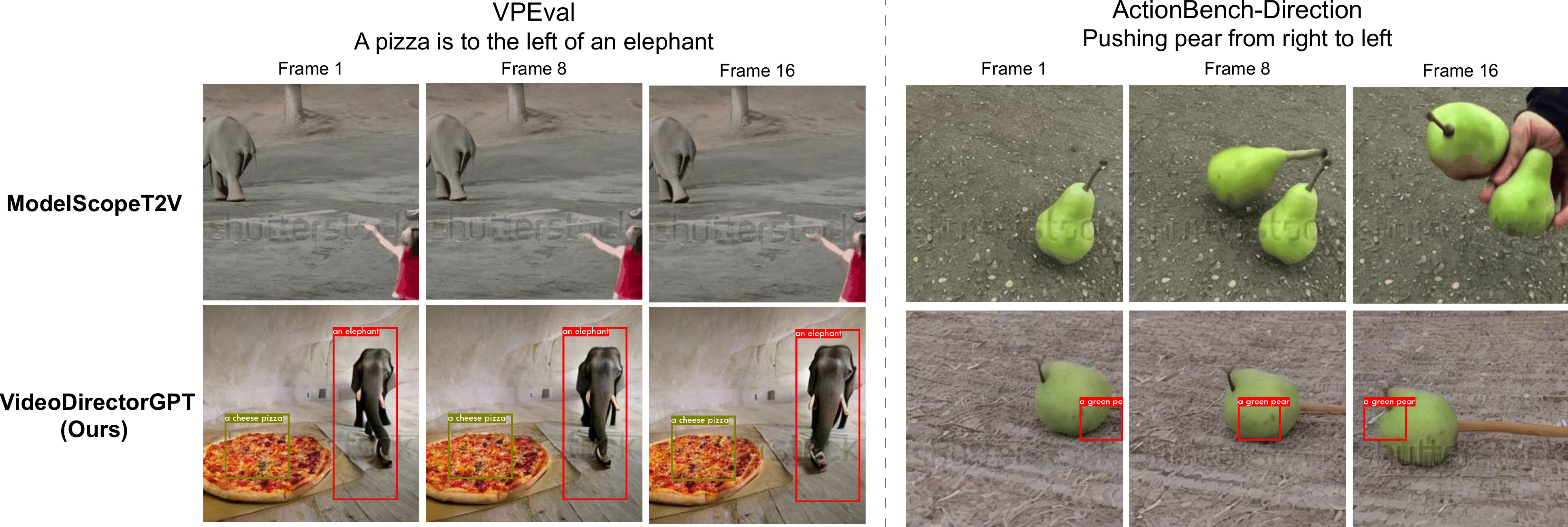}
  \vspace{-5pt}
\caption{
Generated examples on \textbf{VPEval Skill-based} and \textbf{ActionBench-Direction} prompts.
}
\label{fig:vpeval_and_actionbench_outputs_main} 
\end{figure}

\noindent\textbf{Layout control results (VPEval Skill-based prompts).}
\Cref{tab:vpeval_actionbench} (left) displays the VPEval accuracy on the VPEval Skill-based prompts.
Our \method{} significantly outperforms \modelscope{} as well as recent T2V/I2V generation models including AnimateDiff~\citep{guo2023animatediff}, I2VGen-XL~\citep{zhang2023i2vgen}, and SVD~\citep{blattmann2023stable} on all layout control skills. 
These results suggest that layouts
generated by our LLM are highly accurate and greatly improve the control of object placements during video generation.
\cref{fig:vpeval_and_actionbench_outputs_main} (left) shows an example where our LLM-generated \videoplan{} successfully guides \videomodule{} to accurately place the objects, while \modelscope{} fails to generate a `pizza'. In \cref{appendix:visualizations}, we show additional examples (see \cref{fig:r1w4}) that our \videoplan{} can generate layouts requiring understanding of physics (\eg{}, gravity, perspectives).

\begin{table}[h]
    \centering
    \vspace{-5pt}
    \resizebox{0.95\linewidth}{!}{
    \begin{tabular}{l ccccc c}
        \toprule
        \multirow{2}{*}{Method}  & \multicolumn{5}{c}{VPEval Skill-based} & ActionBench-Direction \\
        \cmidrule(lr){2-6} \cmidrule(lr){7-7} 
        & Object & Count & Spatial & Scale & Overall Acc. (\%) & Movement Direction Acc. (\%) \\
        \midrule
        \modelscope{} & 89.8 & 38.8 & 18.0 & 15.8 & 40.8 & 30.5 \\
        AnimateDiff & 96.7 & 52.7 & 22.5 & 15.5 & 46.8 & 29.0 \\
        I2VGen-XL & 96.5 & 62.0 & 35.2 & 23.7 & 54.3 & 35.2 \\
        SVD & 93.1 & 46.7 & 29.2 & 15.0 & 45.7 & 30.7 \\
        \midrule
        \rowcolor{lightblue}
        \method{} & \textbf{97.1} & \textbf{77.4} & \textbf{61.1} & \textbf{47.0} & \textbf{70.6} & \textbf{46.5} \\
        \bottomrule
    \end{tabular}
    }
    \caption{
    Single-scene video generation evaluation on layout control (\textbf{VPEval Skill-based}) and object movement (\textbf{ActionBench-Direction}) prompts.
    }
    \label{tab:vpeval_actionbench}
    \vspace{-5pt}
\end{table}

\vspace{3pt}
\noindent\textbf{Object movement results (ActionBench-Direction).}
\Cref{tab:vpeval_actionbench} (right) shows the performance on the ActionBench-Direction prompts which evaluate both temporal understanding and spatial layout control.
For each object, we give prompts that include four movement directions: ‘left to right’, ‘right to left’, ‘top to bottom’, and ‘bottom to top’. Therefore, for a random guess, the movement accuracy is 25\%.
Our \method{} outperforms \modelscope{} and other T2V/I2V models in object movement direction accuracy by a large margin, demonstrating that our LLM-generated layouts can improve the accuracy of object dynamics in video generation.
\cref{fig:vpeval_and_actionbench_outputs_main} (right) shows that our LLM-generated \videoplan{} guides the \videomodule{} module to correctly locate and move the `pear' in the accurate direction,
whereas the `pear' in the \modelscope{} video is not moved and just gets grabbed.
In \cref{appendix:visualizations}, we show additional examples (see \cref{fig:r2w2}) where for static objects (\eg{}, a bottle), movement is often depicted via camera motion, whereas for objects that can naturally move (\eg{}, airplane), our \method{} directly shows object movements.

\begin{wraptable}[15]{r}{6.3cm}
\centering
\resizebox{\linewidth}{!}{%
    \begin{tabular}{l ccc}
    \toprule
    \multirow{2}{*}{Method}  & \multicolumn{3}{c}{MSR-VTT} \\
    \cmidrule(lr){2-4}  
    & FVD ($\downarrow$) & FID ($\downarrow$)    & CLIPSIM ($\uparrow$)  \\
    \midrule
    \multicolumn{2}{l}{\textcolor{gray}{Different arch / Training data}} \\
    NUWA & ${-}$ & ${47.68}$  & ${0.2439}$ \\
    CogVideo (Chinese) & ${{-}}$  & ${24.78}$  & ${{0.2614}}$  \\
    CogVideo (English) & ${{1294}}$  & ${23.59}$    & ${{0.2631}}$  \\
    MagicVideo & ${{1290}}$  & ${-}$  & ${{-}}$ \\
    VideoLDM  & ${{-}}$ & ${-}$  & ${{0.2929}}$	\\
    Make-A-Video &  ${{-}}$ & ${13.17}$  & 0.3049 \\
    \midrule
    \multicolumn{3}{l}{\textcolor{gray}{Same video backbone \& Test prompts}} \\
    \modelscope{}$^\dagger$ & $683$ & $12.32$  & \textbf{0.2909} \\
    \rowcolor{lightblue}
    \method{} & \textbf{550} & \textbf{12.22}  & {0.2860} \\
    \bottomrule
    \end{tabular}
}
\caption{
    Comparison results on MSR-VTT.
    \modelscope{}$^\dagger$:
    Our replication
    on the same 2990 random test prompts.
}
\label{table:msrvtt}
\end{wraptable}

\vspace{3pt}
\noindent\textbf{Open-domain results.}
\Cref{table:msrvtt} presents the FVD, FID, and CLIPSIM scores on MSR-VTT. 
It's worth noting that our model is built on the pretrained \modelscope{} backbone.
Despite this, our method demonstrates good improvements in FVD and competitive scores in FID and CLIPSIM.
In addition, our method achieves better or comparable performance to models trained with larger video data (e.g., Make-A-Video) or with higher resolution (e.g., VideoLDM). 
Evaluation results on UCF-101 are presented in \cref{appendix:additional_experiments}, where we achieve significant
improvement in FVD (\modelscope{} 1093 vs. Ours 748) and similar performance in the Inception Score (\modelscope{} 19.49 vs. Ours 19.42).

\subsection{Multi-Scene Video Generation}\label{sec:multi_scene_video_gen}

\begin{figure}[t]
\centering
    \includegraphics[width=.95\linewidth]{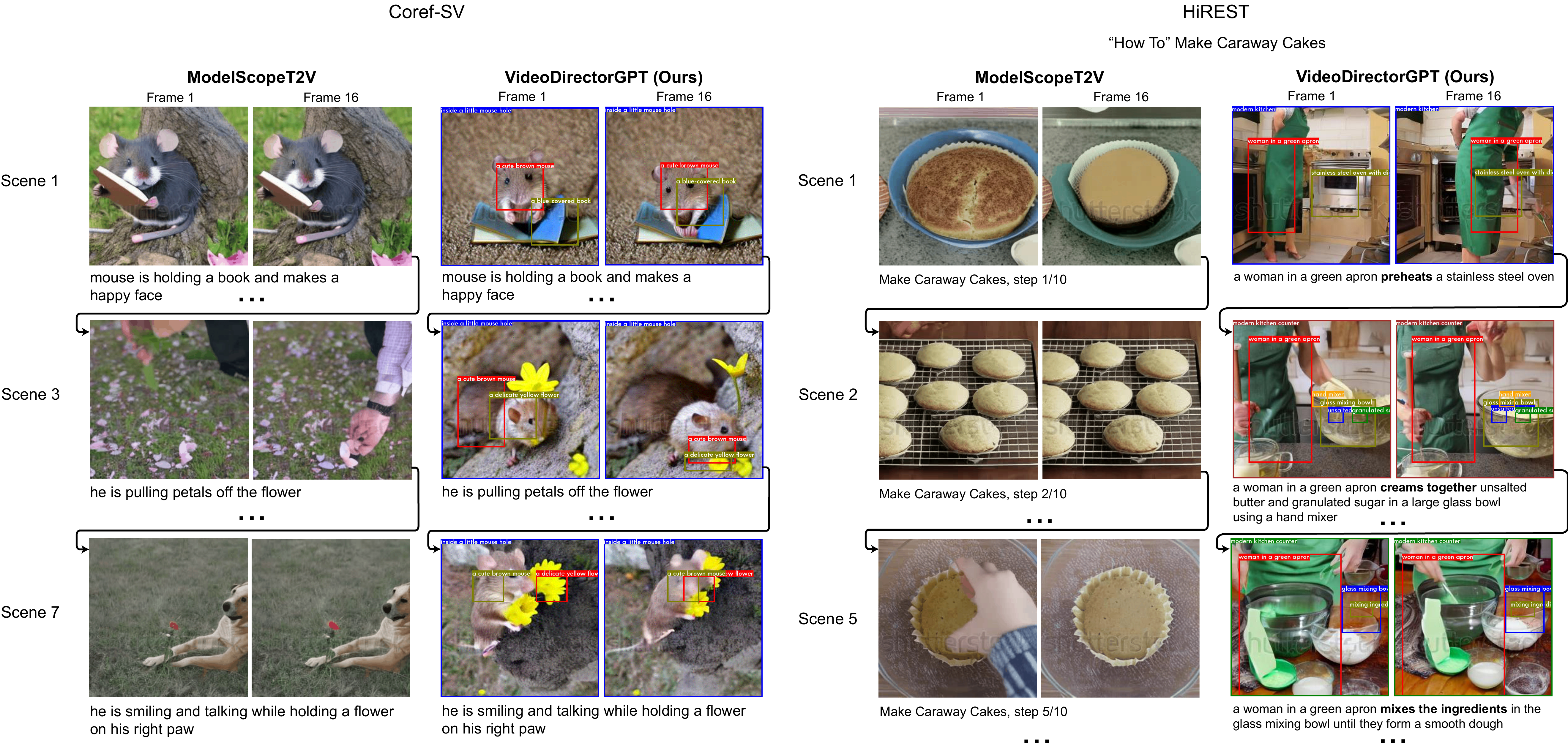}
    \caption{
        Generation examples on \textbf{\corefdata{} (left)} and \textbf{HiREST (right)}.
        Our \method{} is able to generate detailed \textit{video plans} and visually consistent videos. 
    }
\label{fig:pororo_and_hirest_outputs}
\end{figure}

\noindent\textbf{Multiple sentences to multi-scene videos generation (ActivityNet Captions / Coref-SV).}
As shown in the left two blocks of \Cref{tab:multi_scene_video},
our \method{} outperforms \modelscope{} in visual quality (FVD/FID) and multi-scene object temporal consistency.
Notably, for \corefdata{}, our \method{} achieves higher object consistency than \modelscope{} even with GT co-reference 
(where pronouns are replaced with their original noun counterparts, acting as oracle information; e.g., ``she picked up ...'' becomes ``cat picked up ...'').
{As we do not have ground-truth videos for \corefdata{}, we only use it to evaluate consistency when co-references are used across scenes.}
\cref{fig:pororo_and_hirest_outputs} (left) shows a video generation example from \corefdata{},
where the LLM-generated \videoplan{} can guide the \videomodule{} module to generate the same mouse across scenes consistently,
whereas \modelscope{} generates a hand and a dog instead of a mouse in later scenes.

\begin{table}[t]
\centering
\resizebox{.99\linewidth}{!}{
\begin{tabular}{l ccc c cc}
\toprule
\multirow{2}{*}{Method}  & \multicolumn{3}{c}{ActivityNet Captions} & Coref-SV & \multicolumn{2}{c}{HiREST} \\
\cmidrule(lr){2-4} \cmidrule(lr){5-5} \cmidrule(lr){6-7}
& FVD ($\downarrow$) & FID ($\downarrow$) & Consistency ($\uparrow$) &  Consistency ($\uparrow$) & FVD ($\downarrow$) & FID ($\downarrow$) \\
\midrule
\modelscope{} & 980 & 18.12 & 46.0 & 16.3 & 1322 & 23.79\\
\modelscope{} (with GT co-reference; oracle) & -& -& -& 37.9 & - & - \\
\modelscope{} (with prompts from our \videoplan{}) & -& -& -& - & 918 & 18.96 \\
\rowcolor{lightblue}
\method{} (Ours) & \textbf{805} & \textbf{16.50} & \textbf{64.8} & \textbf{42.8} & \textbf{733} & \textbf{18.54} \\
\bottomrule
\end{tabular}
}
\caption{
    Multi-scene video generation with multiple input text (ActivityNet Captions and \corefdata{})
    and single sentence (HiREST).
    \textit{GT co-reference}: replacing co-reference pronouns in \corefdata{} with the original object names (e.g., ``his friends'' may become ``dog's friends''). 
    \textit{Prompts from our \videoplan{}}: 
    an enhanced benchmark for \modelscope{} using scene descriptions from our \videoplan{}.
}
\label{tab:multi_scene_video}
\end{table}

\vspace{5pt}
\noindent\textbf{Single sentences to multi-scene videos generation (HiREST).}
The right block of \Cref{tab:multi_scene_video} shows
our \method{} achieves better visual quality scores (FVD/FID) than \modelscope{} on the HiREST dataset.
Moreover, by comparing with an enhanced \modelscope{} benchmark, where videos are generated from each scene description in our GPT-4-produced video plan, we demonstrate the effectiveness of prompts generated by GPT-4.
However, it still performs worse than our \method{}, also showing the effectiveness of our \videomodule{}.
As shown in \cref{fig:pororo_and_hirest_outputs} (right), our LLM can generate a step-by-step \videoplan{} from a single prompt and our \videomodule{} can generate consistent videos following the plan, 
showing how to make caraway cakes (a British seed cake). In contrast, \modelscope{} repeatedly generates the final caraway cake (which is visually inconsistent). 
An additional example with human (see \cref{fig:multi_scene_boy}) is shown in \cref{appendix:visualizations}.

\begin{figure}[t]
\centering
    \vspace{-2mm}
    \includegraphics[width=.85\linewidth]{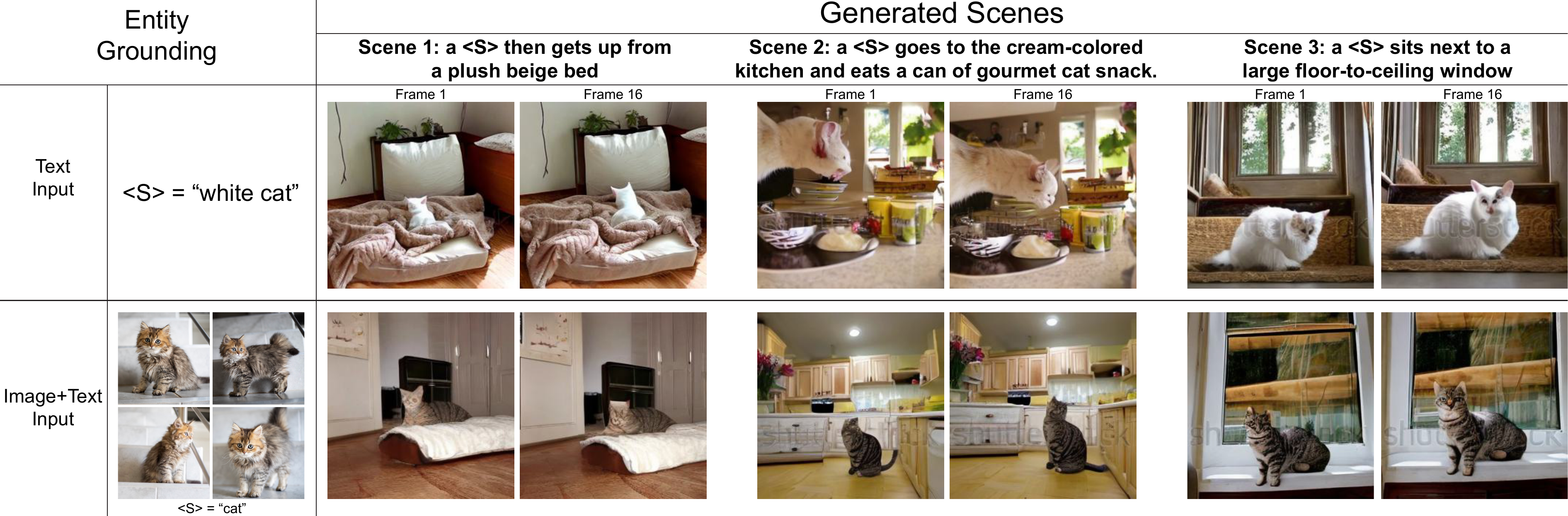} 
    \vspace{-3mm}
    \caption{
    Video generation with text-only (top) and image+text (bottom) inputs with \method{}.
    By giving text or image+text inputs,
    users can generate images where 
    the identities of the entities are preserved across multiple scenes.
    }
\label{fig:image_input_main}
\end{figure}

\subsection{Additional Analysis}\label{sec:additional_analysis}

\noindent\textbf{Generating videos with custom image exemplars.}
Our \videomodule{} can obtain CLIP image embeddings either from user-provided image exemplars or from entity text descriptions via the unCLIP Prior.
In \cref{fig:image_input_main}, we demonstrate that our \videomodule{} can flexibly take either text-only or image+text descriptions as input to generate consistent multi-scene videos.

\vspace{3pt}
\noindent\textbf{Dynamic layout strength control with GPT-4.}
The number of denoising steps with layout guidance $\alpha$
controls layout control strength of
\videomodule{} (see \cref{appendix:visualizations} for more detailed illustration).
Instead of using a static $\alpha$ value, we explore dynamically adjusting it during the \videoplan{} generation by asking LLM how much layout guidance needs to be used for each prompt.
\Cref{tab:ablation_alpha} compares static $\alpha$ values
and dynamic $\alpha$ values (LLM-Dynamic-$\alpha$), showing that LLMs help achieve a good balance in the quality-layout trade-off.

\begin{table}[t]
    \centering
    \resizebox{0.85\linewidth}{!}{
    \begin{tabular}{l ccc c}
        \toprule
        \multirow{2}{*}{\parbox{4cm}{\# Denoising steps with layout guidance (out of 50)}} & \multicolumn{3}{c}{MSR-VTT} & ActionBench-Direction \\
        \cmidrule(lr){2-4} \cmidrule(lr){5-5} 
        & FVD ($\downarrow$) & FID ($\downarrow$) & CLIPSIM ($\uparrow$) & Move Direction Acc. (\%) \\
        \midrule
        $\alpha=0.1$ (5 steps) & 550 & \textbf{12.22} & \textbf{0.2860} & 46.5  \\
        $\alpha=0.2$ (10 steps) & 588 & 17.25 & 0.2700 & \textbf{59.8} \\
        $\alpha=0.3$ (15 steps) & 593 & 17.17 & 0.2702 & 57.8 \\
        LLM-Dynamic-$\alpha$ (5-15 steps) & \textbf{523} & 13.75 & 0.2790 & 56.8 \\
        \bottomrule
    \end{tabular}
    }
    \caption{Ablation of the denoising steps with layout guidance (via guided 2D attentions) in MSR-VTT and ActionBench-Direction prompts, where
    $\alpha=\frac{\text{\# steps with layout guidance}}{\text{\# total steps}}$.
    }
    \label{tab:ablation_alpha}
\end{table}

\noindent\textbf{Video Planner LLMs: GPT-4, GPT-3.5-turbo, and LLaMA2.}
We conducted an ablation study using the open-sourced LLaMA2-7B and LLaMA2-13B models~\citep{touvron2023llama} as well as GPT-3.5-turbo in the video planning stage on MSR-VTT and ActionBench-Directions. 
We evaluated LLaMA2 in both fine-tuned and zero-shot settings. To collect data for fine-tuning LLaMA2, we randomly sampled 2000 prompts from WebVid-10M~\citep{bain2021frozen} and generated layouts with GPT-4. In addition to the visual quality (FID/FVD) and video-text alignment (CLIPSIM) scores on MSR-VTT and movement direction accuracy score on ActionBench-Directions, we also report the number of samples whose output layouts can be successfully parsed (\eg, \#Samples). Evaluation setup details are given in \cref{sec:ablation_study}.
\Cref{table:ablation_llms} shows that zero-shot LLaMA2-13B struggles to generate layouts that can be successfully parsed (\eg, only 82 out of 600). It also achieves worse scores on all three metrics (FID, FVD, and CLIPSIM) on MSR-VTT, as well as the movement direction accuracy on ActionBench-Directions, which shows that GPT models have stronger in-context learning skills and can generate more reasonable layouts.
On the other hand, all scores can be improved by fine-tuning on layouts generated by GPT-4.
Fine-tuning open-sourced LLMs can be a future direction worthwhile to explore.
Another point to note is that GPT-3.5-turbo and GPT-4 achieve very similar performance on MSR-VTT, but GPT-3.5-turbo performs significantly worse than GPT-4 on ActionBench-Directions. This suggests that GPT-3.5-turbo is less effective at handling prompts requiring strong layout control compared to GPT-4.

\noindent\textbf{Challenging cases.} 
The bounding box-based control of our method poses two main challenges.
Firstly, some objects might not follow the bounding box control well when there are too many overlapping bounding boxes. For example, in \cref{fig:pororo_and_hirest_outputs} right, the ``hand mixer'' and ``unsalted butter'', which have relatively small bounding box sizes, do not follow the box control well. 
Secondly, the control of whether the background is static or moving is usually determined by the prior knowledge of the video generation model. 
For example, as shown in \cref{fig:r2w2}, it’s hard for  Layout2Vid to generate a moving bottle with a static background and a moving boat with a completely static background
(see paragraph ``Object movement prompts with different types of objects'' in \cref{appendix:additional_qualitative_examples} for details).

\begin{table}[t]
    \centering
    \resizebox{0.99\linewidth}{!}{
    \begin{tabular}{l cccc cc}
        \toprule
        \multirow{2}{*}{\parbox{3.5cm}{LLMs}} & \multicolumn{4}{c}{MSR-VTT} & \multicolumn{2}{c}{ActionBench-Directions} \\
        \cmidrule(lr){2-5} \cmidrule(lr){6-7} 
        & FVD ($\downarrow$) & FID ($\downarrow$) & CLIPSIM ($\uparrow$) &\#Samples ($\uparrow$) & Movement Direction Acc (\%)  &\#Samples ($\uparrow$) \\
        \midrule
        \emph{Fine-tuned} \\
        LLaMA2-7B & 580  & 12.69  & 0.2851  & 2749 & 35.8 & 537 \\
        LLaMA2-13B & 556  & 12.40  & 0.2854  & 2786  & 53.3 & 561\\
        \midrule
        \emph{Zero-shot} \\
        LLaMA2-13B & 573 & 13.47 & 0.2792 & 2236 & - & 82 \\
        GPT-3.5-turbo & 558 & 12.27 & 0.2852 & \textbf{2990}  & 49.0 & \textbf{600}\\
        \rowcolor{lightblue}
        GPT-4 (default) & \textbf{550} & \textbf{12.22} & \textbf{0.2860} & \textbf{2990}   & \textbf{59.8} & \textbf{600}\\
        \bottomrule
    \end{tabular}
    }
    \caption{Ablation of video generation with video plans generated from different LLMs on MSR-VTT and ActionBench-Directions.
    }
    \label{table:ablation_llms}
\vspace{-3pt}
\end{table}
\normalsize

\subsection{Human Evaluation}
\label{sec:human_evaluation}

\begin{wraptable}[7]{r}{7cm}
\vspace{-30pt}
\centering
\resizebox{\linewidth}{!}{
    \begin{tabular}{l ccc}
        \toprule
        \multirow{2}{*}{Evaluation category} & \multicolumn{3}{c}{Human Preference (\%) $\uparrow$}\\
        \cmidrule(lr){2-4}
        & Ours
        & \modelscope{} & Tie  \\
        \midrule
        Quality & \textbf{50} & 34 & 16 \\
        Text-Video Alignment & \textbf{58} & 36 & 6 \\
        Object Consistency & \textbf{62} & 28 & 10 \\
        \bottomrule
    \end{tabular}
    }
\caption{Human preference on multi-scene videos generated with
\corefdata{} prompts.
}
\label{table:human_eval}
\end{wraptable}

\textbf{Human preference.}
We conduct a human evaluation study on the multi-scene videos generated by both our \method{} and \modelscope{} on the \corefdata{} dataset.
We show 50 videos to 10 crowd-annotators from AMT\footnote{Amazon Mechanical Turk: \url{https://www.mturk.com}} to rate each video (56 annotators in total, 10 annotators per video) and calculate human preferences for each video with average ratings.
\Cref{table:human_eval} shows that the multi-scene videos (with \corefdata{} prompts) generated by our \method{} are preferred
by human annotators
than \modelscope{} videos in 
all three categories (Quality, Text-Video Alignment, and Object Consistency).

\textbf{Step-by-step error analysis.}
In our error analysis of video planning and generation stages,
we measure 1-5 Likert scale accuracy of four intermediate generation steps: scene descriptions, layouts, consistency groupings, and final video.
While the first three planning steps achieve high scores ($\ge 4.52$),
there is a big score drop in the layout-guided video generation ($4.52\xrightarrow{}3.61$). 
This suggests that our \method{} could generate more accurate videos,
once we have access to a stronger T2V backbone than \modelscope{}.
See \cref{appendix:human_eval} for more detailed analysis.

\section{Conclusion}
\label{sec:conclusion}
In this work, we propose \method{}, a novel 
framework for consistent multi-scene video generation, leveraging the knowledge of LLMs for video content planning and grounded video generation.
In the first stage, we employ GPT-4 as a video planner to craft a \videoplan{}, a multi-component script for videos with multiple scenes.
In the second stage, we use \videomodule{}, a grounded video generation module, to generate videos with layout and consistency control.
Experiments demonstrate that our proposed framework
substantially improves object layout and movement control over state-of-the-art methods on open-domain single-scene video generation, and 
can generate consistent multi-scene videos while maintaining visual quality.

\section*{Acknowledgments}
We thank the reviewers for the thoughtful discussion and feedback. 
This work was supported by ARO W911NF2110220, DARPA MCS N66001-19-2-4031, DARPA KAIROS
Grant FA8750-19-2-1004, NSF-AI Engage Institute DRL211263, ONR N00014-23-1-2356, and Accelerate
Foundation Models Research program. The views, opinions, and/or findings contained in this article are
those of the authors and not of the funding agency.

\bibliography{main}
\bibliographystyle{colm2024_conference}

\newpage

{
\begin{center}
    \textbf{{{\Large Appendix}}}
\end{center}
}

\renewcommand\thesection{\Alph{section}}
\renewcommand\thesubsection{\Alph{section}.\arabic{subsection}}

\appendix
\startcontents[sections]
\printcontents[sections]{l}{1}{\setcounter{tocdepth}{2}}

\vspace{20pt}

\section{Reproducibility Statement}\label{appendix:code}

Our model is built upon the publicly available code repository from GLIGEN~\citep{li2023gligen}\footnote{GLIGEN: \url{https://github.com/gligen/GLIGEN}} and \modelscope{}~\citep{wang2023modelscope}\footnote{ModelScopeT2V: \url{https://modelscope.cn/models/damo/text-to-video-synthesis/summary}, \url{https://github.com/ExponentialML/Text-To-Video-Finetuning/tree/main}}. 
Please see \cref{sec:grounded_video_generation}/\cref{appendix:layout2vid} for model architecture details, \cref{sec:experiments}/\cref{sec:single_scene_evaluation_dataset_appendix}/\cref{sec:multi_scene_evaluation_dataset_appendix} for dataset details.
\textbf{We provide code and video samples (\texttt{video\_samples.html}) in the supplementary material.}

\section{Additional Related Works}\label{appendix:related_works}
\paragraph{Text-to-video generation.}
The text-to-video (T2V) generation task is to generate videos from text descriptions.
Early T2V generation models~\citep{li2017video, li2018storygan} used variational autoencoders (VAE)~\citep{kingma2013auto} 
and generative adversarial networks (GAN)~\citep{goodfellow2020generative},
while multimodal language models~\citep{hong2022cogvideo,wu2022nuwa,villegas2022phenaki,maharana2022storydall,ge2022long,wu2021godiva} and denoising diffusion models~\citep{ho2022imagen,singer2022make,blattmann2023align,khachatryan2023text2video,wang2023gen,yin2023nuwa} have become popular for recent works.
Since training a T2V generation model from scratch is computationally expensive,
recent work often leverages pre-trained text-to-image (T2I) generation models such as Stable Diffusion~\citep{rombach2022high} by finetuning them on text-video pairs~\citep{wang2023modelscope, blattmann2023align}.
While this warm-start strategy enables high-resolution video generation, it comes with the limitation of only being able to generate short video clips, as T2I models lack the ability to maintain consistency through long videos. 
Recent works on long video generation \citep{blattmann2023align, yin2023nuwa, villegas2022phenaki, he2023animate} aim at generating long videos of a few minutes.
However, the generated videos often display the continuation or repetitive patterns of a single action (\eg{}, driving a car) instead of transitions and dynamics of multiple changing actions/events (\eg{}, five steps about how to bake a cake).
In this work, we address this problem of multi-scene video generation with a two-stage framework: using an LLM (\eg{}, GPT-4) to generate a structured \videoplan{} (consisting of stepwise scene descriptions, entities and their layouts) and generating videos using \videomodule{}, a layout-guided text-to-video generation model with consistency control.
Our \videomodule{} infuses layout control and multi-scene temporal consistency into a pretrained T2V generation model via data and parameter-efficient training, while preserving its original visual quality.

\section{Video Planning}\label{appendix:prompt_templates}

\subsection{GPT-4 Prompt Templates}
\begin{figure}[h]
\begin{center}
    \includegraphics[width=0.99\linewidth]{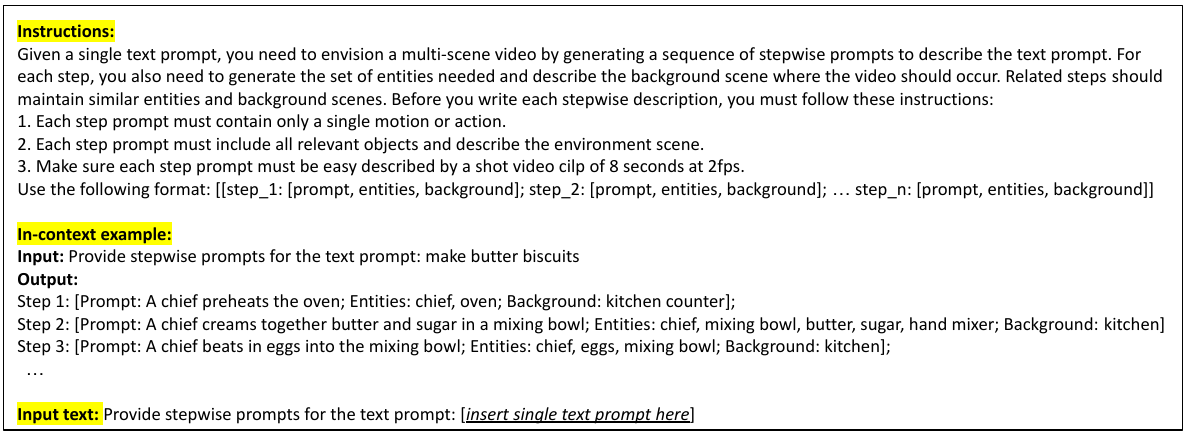}
    \caption{Prompt template for the 1st video planning step (scene descriptions and entities/background generation).}
\label{fig:steps_prompt}
\vspace{-4mm}
\end{center}
\end{figure}

\begin{figure}[h]
\begin{center}
    \includegraphics[width=0.99\linewidth]{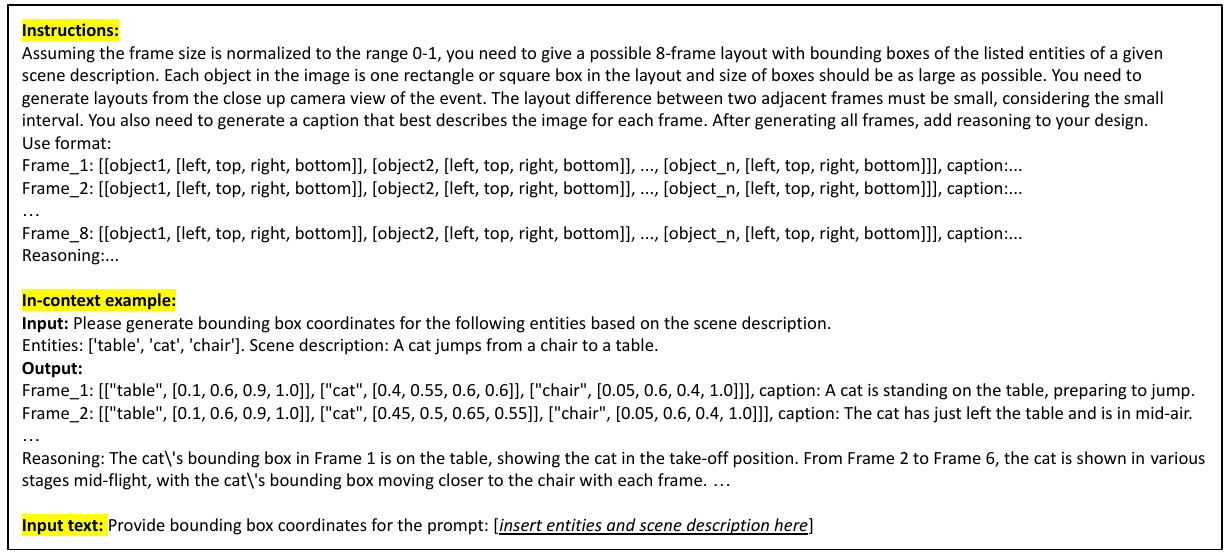}
    \caption{Prompt template for the 2nd video planning step (layout generation).}
\label{fig:layout_prompt}
\vspace{-4mm}
\end{center}
\end{figure}

In this section, we provide the prompt templates we give to our video planner (\cref{sec:grounded_video_generation}).
The \videoplan{} is generated in two steps by prompting GPT-4\footnote{We employ the \texttt{gpt-4-0613} version.} with different in-context examples (we use 1 and 5 in-context examples for the first and second steps, respectively).
In the first step (see \cref{fig:steps_prompt}), we ask GPT-4 to expand a single text prompt into a multi-scene \videoplan{}. Each scene comes with a text description, a list of entities, and a background.
In the second step (see \cref{fig:layout_prompt}), we generate a list of bounding boxes for the entities
in each frame based on the list of entities and the scene description. 
In line with VPGen~\citep{cho2023visual}, we utilize the $[x_0, y_0, x_1, y_1]$ format for bounding boxes, where each coordinate is normalized to fall within the range [0,1].
For in-context examples, we present 0.05 as the minimum unit for the bounding box, equivalent to a 20-bin quantization over the [0,1] range.

\subsection{API Cost}
Using GPT-4 tokenizer, the average input/output token lengths of each step are 2K/1K for the first step and 6K/1K for the second step.
Using GPT-4, it takes 0.12 USD and 0.24 USD for the inference of the first and second steps, respectively.

\section{\videomodule{}}\label{appendix:layout2vid}

\subsection{Preliminaries: T2V Generation Backbone}\label{appendix:layout2vid_preliminaries}

We implement \videomodule{} by injecting layout control capability into \modelscope{}~\citep{wang2023modelscope}, a public text-to-video generation model based on Stable Diffusion~\citep{rombach2022high}.
\modelscope{} consists of
(1) a CLIP ViT-H/14~\citep{radford2021learning} text encoder,
(2) an autoencoder, and
(3) a diffusion UNet~\citep{ronneberger2015unet,ho2020denoising}.
Given a $T$ frame video $\bm{x}\in\mathbb{R}^{T\times 3 \times H \times W}$ with video caption $\bm{c}$ and frame-wise layouts $\{\bm{e}\}_{i=1}^T$, \modelscope{} first uses an autoencoder
to encode the video into a latent representation $\bm{z}=\mathcal{E}(\bm{x})$.
The diffusion UNet performs denoising steps in the latent space to generate videos, conditioned on the CLIP text encoder representation of video captions.
The UNet comprises a series of spatio-temporal blocks, each containing four modules: spatial convolution, temporal convolution, spatial attention, and temporal attention. Since the original \modelscope{} does not offer control beyond the text input, we build upon the 2D attention module in the spatial attention module to create `Guided 2D Attention', which allows for spatial control using bounding boxes.

\begin{figure}[t]
\centering
    \includegraphics[width=.65\linewidth]{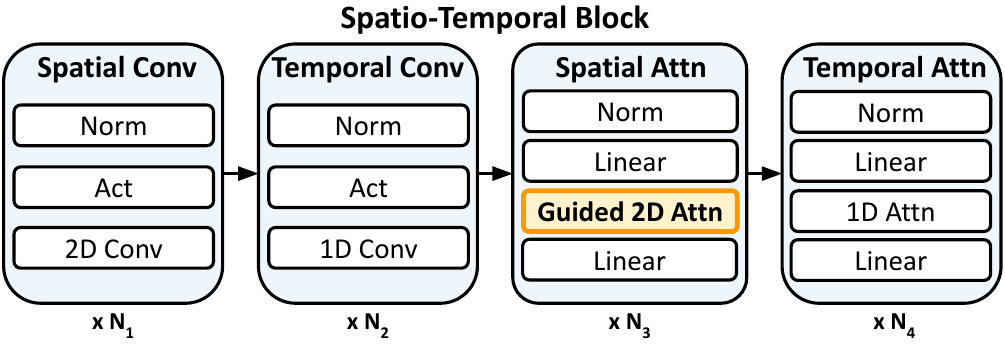}
    \vspace{-3mm}
    \caption{
    \textbf{Spatio-temporal blocks within the diffusion UNet of \textbf{\videomodule{}}}.
    The spatio-temporal block comprises four modules: spatial convolution, temporal convolution, spatial attention, and temporal attention. We adopt settings from \modelscope{}, where (N1, N2, N3, N4) are set to (2, 4, 2, 2). We use ``Act'' to represent activation layers.
    }
\label{fig:network}
\end{figure}

\subsection{Layout2Vid Spatio-Temporal Blocks}
\label{appendix:spatio-temporal-blocks}
\cref{fig:network} illustrates the spatio-temporal blocks within the diffusion UNet of \videomodule{}.
The spatio-temporal block comprises four modules: spatial convolution, temporal convolution, spatial attention, and temporal attention.
Following~\cite{li2023gligen}, we build upon the 2D attention to create the guided 2D attention, which enables layout-guided video generation 
with explicit spatial control over a list of entities represented by their bounding boxes, as well as visual and text content.

\subsection{Training and Inference Details}\label{appendix:training_inference_details}

\Cref{table:hyperparameters_gen} contains the model architecture details and training/inference parameter settings for our \videomodule{}.

\paragraph{Training.} The highlight of our \videomodule{} training is that it was conducted solely on image-level data with bounding box annotations.
We trained the MLP layers for grounding tokens and the gated self-attention layers with the same bounding-box annotated data used in GLIGEN~\citep{li2023gligen}, which consists of 0.64M images. 
We train \videomodule{} for 50k steps, which takes only two days with 8 A6000 GPUs (each 48GB memory). 
All the remaining modules in the spatio-temporal block (see main paper Fig. 4) are frozen during training. 
We illustrate the training and inference procedure of \videomodule{} in main paper Fig. 3. 

\paragraph{Inference.} During inference,
we use the Karlo implementation of unCLIP Prior
to the entities to convert the texts into their corresponding CLIP image embeddings, and CLIP text encoder to get their corresponding text embeddings. We use CLIP ViT-L/14 as a backbone during training to be consistent with Karlo. This helps us to preserve the visual consistency of the same object by using the same image embedding across scenes. In addition, we use PLMS~\cite{stablediffusion} (based on PNDM~\cite{liu2022pseudo}) as our default sampler following GLIGEN~\citep{li2023gligen}, since we found no consistent improvements for visual quality and video-text alignment scores when switching to the DDIM~\citep{song2020denoising} sampler used in ModelScopeT2V. The comparison of the PLMS and DDIM samplers is shown in \Cref{table:ablation_samplers}.

\begin{table}[h]
\setlength{\tabcolsep}{4pt}
\centering
\resizebox*{0.72\textwidth}{!}{
    \begin{tabular}{lc@{\hskip 0.75in}lc}
    \toprule
    \textbf{\emph{Architecture}} & & \textbf{\emph{Training}} \\
    $z$-shape & $32\times32\times4$ & \# Train steps & 50k \\
    Channels & 320 & Learning rate & 5e-5\\
    Depth & 2 & Batch size per GPU & 32 \\
    Attention scales & 1,0.5,0.25 & \# GPUs & 8 \\
    Number of heads & 8 & GPU-type & A6000 (48GB) \\
    Head dim & 64 & Loss type & MSE\\
    Channel multiplier & 1.2,4,4 & Optimizer & AdamW\\
    \midrule
    \textbf{\emph{CA Conditioning}} & & \textbf{\emph{Inference}} \\
    Context dimension & 1024 & Sampler & PLMS \\
    Sequence length & 77 & \# denoising steps & 50\\
    \bottomrule
\end{tabular}
}
\caption{Hyperparameters for \videomodule{}.}
\label{table:hyperparameters_gen}
\end{table}

\begin{table}[h]
\begin{center}
    \resizebox{0.5\linewidth}{!}{
    \begin{tabular}{lccc}
        \toprule
        Samplers & FVD ($\downarrow$) & FID ($\downarrow$) & CLIPSIM ($\uparrow$) \\
        \midrule
        DDIM  & \textbf{508} & {12.52} & 0.2848  \\
        PLMS  & {550} & \textbf{12.22} & \textbf{0.2860}  \\
        \bottomrule
    \end{tabular}
    }
\end{center}
\caption{
Comparison between PLMS and DDIM sampler on MSR-VTT.}
\label{table:ablation_samplers}
\end{table}

\section{Human-in-the-Loop Video Plan Editing}
\label{appendix:human_in_the_loop}

\begin{figure}[h]
\begin{center}
    \includegraphics[width=.9\linewidth]{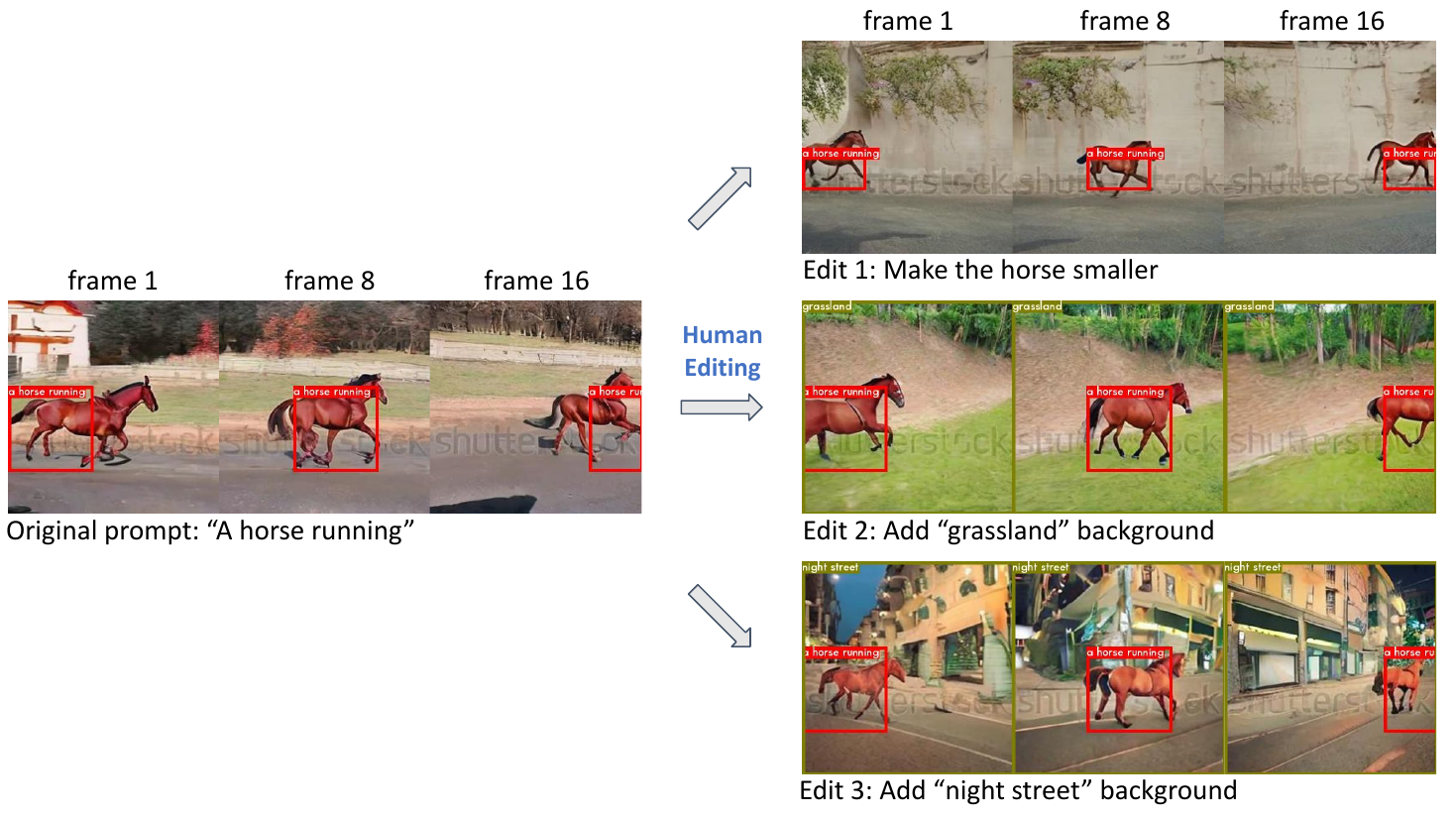}
    \vspace{-2mm}
    \caption{
        Video generation examples for human-in-the-loop editing.
        Users can modify the video plan (\eg, add/delete objects, change the background and entity layouts, \etc{}) to generate customized video contents.
        Given the same text prompt ``A horse running'', we provide visualizations with a smaller horse and different backgrounds (\ie, ``night street'' and ``grassland'').
    }
\label{fig:human_edit}
\end{center}
\end{figure}

One useful property of our \method{} is that it is very flexible to edit \videoplan{}s via human-in-the-loop editing.
Given an LLM-generated \videoplan{}, users can generate customized content by adding/deleting/replacing the entities, adding/changing the background, and modifying the bounding box layouts of entities.
\Cref{fig:human_edit} showcases the example videos generated from the text prompt ``A horse running'', and videos generated from the \videoplan{} modified by users (\ie, changing the object sizes and backgrounds).

\section{Experiment Setup}\label{sec:experiments_appendix}
We provide additional details on our experiment setups (\cref{sec:experiments} in the main paper)
below.

\subsection{Evaluated Models}
\label{sec:baseline_models_appendix}

We compare our \method{} to 6 popular T2V generation models, 
NUWA~\citep{wu2022nuwa1}, 
CogVideo~\citep{hong2022cogvideo},
VideoLDM~\citep{blattmann2023align},
MagicVideo~\citep{zhou2022magicvideo}, 
Make-A-Video~\citep{singer2022make},
and \modelscope{}~\citep{wang2023modelscope}.
Since NUWA, VideoLDM, MagicVideo, Make-A-Video, and CogVideo (English) are not publicly available,
we primarily compare our \method{} with \modelscope{}, and present comparisons with the other models on the datasets for which their papers have provided results.
\modelscope{} is the closest baseline to our framework among all these models, because our \videomodule{} utilizes its frozen weights and only trains a small set of new parameters to add spatial control and temporal consistency across multiple scenes.

\subsection{Prompts for Single-Scene Video Generation}
\label{sec:single_scene_evaluation_dataset_appendix}

For single-scene video generation,
we conduct experiments with
VPEval Skill-based prompts to evaluate layout control~\citep{cho2023visual},
ActionBench-Direction prompts to assess object dynamics~\citep{wang2023paxion},
and MSR-VTT to cover diverse open-domain scenes~\citep{xu2016msr-vtt}.

\textbf{VPEval Skill-based prompts} 
evaluate different object-centric layout control skills in text-to-image/video generation.
We randomly sample 100 prompts for each of the four skills:
Object (generation of a single object),
Count (generation of a specific number of objects),
Spatial (generation of two objects with a spatial relation; \eg{}, left/right/above/below),
and Scale (generation of two objects with a relative scale relation; \eg{}, bigger/smaller/same).

\textbf{ActionBench-Direction prompts} 
evaluate the action dynamics (object movement directions) in video language models.
We prepare the prompts by sampling video captions from ActionBench-SSV2~\citep{wang2023paxion} and balancing the distribution of movement directions.
Concretely, we select captions from the ActionBench-SSV2 validation split that include phrases like `right to left' or `left to right' (\eg{}, `pushing a glass from left to right'), which are common phrases describing movement directions in the captions.
Then we augment these prompts by switching the directions to each of four directions: `left to right', `right to left', `top to bottom', and `bottom to top' to create 100 prompts for each direction.
We call the resulting 400 prompts as \textit{ActionBench-Direction} prompts. These prompts ensure a balanced distribution of movement directions while maintaining diversity in objects.

\textbf{MSR-VTT} is an open-domain video captioning dataset, which allows us to check if 
our \videomodule{} maintains the original visual quality and text-video alignment performance of the \modelscope{} backbone after integration of the layout/movement control capabilities.
The MSR-VTT test set comprises 2,990 videos, each paired with 20 captions.
Following VideoLDM~\citep{blattmann2023align}, we sample one caption from the 20 available captions for each video and use the 2,990 corresponding generated videos for evaluation.

\subsection{Prompts for Multi-Scene Video Generation}
\label{sec:multi_scene_evaluation_dataset_appendix}

For multi-scene video generation,
we experiment with two types of input prompts:
(1) a list of sentences describing events -- ActivityNet Captions~\citep{krishna2017dense} and
\corefdata{} prompts based on Pororo-SV~\citep{li2018storygan}
and
(2) a single sentence from which models generate multi-scene videos  -- HiREST~\citep{zala2023hierarchical}.

\textbf{ActivityNet Captions} is a dense-captioning dataset designed for detecting and describing multiple events in videos using natural language.
For the multi-scene video generation task,
we use 165 randomly sampled videos from the validation split
and use the event captions as input for \modelscope{} and our \method{}.
When calculating object consistency, we find the subject of the first event caption via spaCy dependency parser~\citep{spacy2} and check its appearance in multiple scenes.

\textbf{\corefdata{}}
is a new multi-scene text description dataset
that we propose to 
evaluate the consistency of object appearances across multi-scene videos.
We prepare the \corefdata{} prompts by augmenting the Pororo-SV dataset~\citep{li2018storygan,PororoQA2017Kim},
which consists of multi-scene paragraphs from the ``Pororo the Little Penguin'' animated series.
To evaluate the temporal consistency of video generation models trained on real-world videos, we extend its original animation characters (\eg{}, Pororo) to humans and common 
animals and examine their appearance across different scenes.
Concretely, we sample 10 episodes, each consisting of multiple scenes (6.2 scenes per episode on average).
Then, we replace the first appearance of a character with one of the predefined 10 real-world entities (\eg{}, person/dog, etc.) and replace the remaining appearances of the character with pronouns (\eg{}, he/she/it/etc.).
In total, we obtain 100 episodes (=10 episodes $\times$ 10 entities) in \corefdata{}.
In order to generate visually consistent entities,
the multi-scene video generation models would need to address the co-reference of these target entities across different scenes.
We use the final scene descriptions as input for both \modelscope{} and \method{}. When calculating object consistency, we use the selected 
entity as the target object.

\textbf{HiREST} provides step annotations for instructional videos paired with diverse `How to' prompts (\eg{}, a video paired with `how to make butter biscuits' prompt is broken down into a sequence of short video clips of consecutive step-by-step instructions).
For the multi-scene video generation task, we employ 175 prompts from the test splits,
where we only include the prompts with step annotations, to ensure that it is possible to create multi-scene videos from the prompts.
Note that instead of providing a list of scene description sentences like in ActivityNet Captions/\corefdata{}, we only give the single high-level 
`How to' prompt and let the models generate a multi-scene video from it.
In \method{}, our LLM can automatically 
generate the multi-scene \videoplan{} and video from the input prompt.
In contrast, for the \modelscope{} baseline, we help the model understand the different number of scenes to generate by
pre-defining the number of scenes $N$, and independently generate $N$ videos by appending the suffix ``step $n/N$'' to the prompt for $n$-th scene (\eg{}, ``Cook Beet Greens, step 1/10'').
To ensure that our \method{} videos and \modelscope{} videos are equal in length,
we use the same number of scenes generated by our LLM during the planning stage for \modelscope{}. As mentioned in main paper \cref{sec:multi_scene_video_gen}, we also try an ablation with giving \modelscope{} the LLM generated prompts instead of just ``step $n/N$''.

\section{Human Evaluation Details}
\label{appendix:human_eval}

We provide details of our human evaluation and error analysis setups, as well as the error analysis results described in \cref{sec:experiments}.

\paragraph{Human evaluation details.}
We conduct a human evaluation study on the multi-scene videos generated by both our \method{} and \modelscope{} on the \corefdata{} dataset.
Since we know the target entity and its co-reference pronouns in the \corefdata{} prompts,
we can compare the temporal consistency of the target entities across scenes.
We evaluate the human preference between videos from two models in each category of Quality, Text-Video Alignment, and Object Consistency:
\begin{itemize}
    \item \textit{Quality:} it measures how well the video looks visually.
    \item \textit{Text-Video Alignment:} it assesses how accurately the video adheres to the input sentences.
    \item \textit{Object Consistency:} it evaluates how well the target object maintains its visual consistency across scenes.
\end{itemize}
We show 50 videos to ten crowd-annotators from AMT\footnote{Amazon Mechanical Turk: \url{https://www.mturk.com}} to rate each video (28 unique annotators, ten annotators rate each video) and calculate human preferences for each video with average ratings. All videos are randomly shuffled such that annotators do not know which model generated each video.
To ensure high-quality annotations, we require they have an AMT Masters, have completed over 1000 HITs, have a greater than 95\% approval rating, and are from one of the United States, Great Britain, Australia, or Canada (as our task is written in the English language). We pay annotators \$0.10 to evaluate a video (roughly \$12-14/hr).

\paragraph{Step-by-step error analysis details.}
We conduct an error analysis on each step of our single sentence to multi-scene video generation pipeline for HiREST prompts. We analyze the generated multi-scene text descriptions,
layouts, and entity/background consistency groupings to evaluate our video planning stage and the final video to evaluate the video generation stage.
\begin{itemize}
    \item \textit{Multi-Scene Text Descriptions Accuracy}: we measure how well these descriptions depict the intended scene from the original prompt (\eg{}, if the original prompt is ``Make buttermilk biscuits'' the descriptions should describe the biscuit-making process and not the process for pancakes).
    \item \textit{Layout Accuracy}: we measure how well the generated layouts showcase a scene for the given multi-scene text description (\eg{}, the bounding boxes of ingredients should go into a bowl, pan, etc. instead of randomly moving across the scene).
    \item \textit{Entity/Background Consistency Groupings Accuracy}: we measure how well the generated entities and backgrounds are grouped (\eg{}, entities/backgrounds that should look consistent throughout the scenes should be grouped together).
    \item \textit{Final Video Accuracy}: we measure how well the generated video for each scene matches the multi-scene text description (\eg{}, if the multi-scene text description is ``preheating an oven'', the video should accurately reflect this).
\end{itemize}
We ask an expert annotator to rank the generations (multi-scene text description, layouts, etc.) on a Likert scale of 1-5 for 50 prompts/videos. Analyzing the errors at each step enables us to check which parts are reliable and which parts need improvement. As a single prompt/video can contain many scenes, to simplify the process for layout and final video evaluation of a prompt/video, we sub-sample three scene layouts and corresponding scene videos and average their scores to obtain the ``Layout Accuracy'' and ``Final Video Accuracy.''

\paragraph{Step-by-step error analysis results.}
As shown in \Cref{table:error_analysis}, our LLM-guided planning scores high accuracy on all three components (up to 4.52), whereas the biggest score drop happens in the layout-guided video generation (4.52 $\rightarrow$ 3.61).
This suggests that our \method{} could generate even more accurate videos,
once we have access to a stronger T2V backbone than \modelscope{}.

\begin{table}[h]
\begin{center}
\resizebox{0.99\linewidth}{!}{
    \begin{tabular}{ccc c}
        \toprule
        \multicolumn{3}{c}{Stage 1: Video Planning (with GPT-4)} & Stage 2: Video Generation (with \videomodule{}) \\
        \cmidrule(lr){0-2} \cmidrule(lr){4-4} 
        Multi-scene Text Descriptions ($\uparrow$) & Layouts ($\uparrow$) & Entity/Background Consistency Groupings  ($\uparrow$) & Final Video ($\uparrow$) \\
        \midrule
        4.92 & 4.69 & 4.52 & 3.61 \\
        \bottomrule
    \end{tabular}
}
\end{center}
\caption{Step-wise error analysis of \method{} video generation pipeline on HiREST prompts.
We use a Likert scale (1-5) to rate the accuracy of the generated components at each step.
}
\label{table:error_analysis}
\end{table}

\section{Ablation Studies}
\label{sec:ablation_study}

In this section, we provide ablation studies on our design choices,
including using different LLMs for video planning, the number of layout-guided denoising steps,
different embeddings for layout groundings, and layout representation formats.

\subsection{Video Planner LLMs: GPT-4, GPT-3.5-turbo, and LLaMA2}

We conducted an ablation study using the open-sourced LLaMA2-7B and LLaMA2-13B models~\citep{touvron2023llama} as well as GPT-3.5-turbo in the video planning stage on MSR-VTT and ActionBench-Directions. 

\paragraph{Evaluation settings.} In this study, we evaluated LLaMA2 in both fine-tuned and zero-shot settings. For the zero-shot evaluation, we used the same number of in-context examples (\ie, 5 examples) for both GPT-3.5-turbo and GPT-4, but only 3 in-context examples for the LLaMA2 models due to their 4096 token length limits. To collect data for fine-tuning LLaMA2, we randomly sampled 2000 prompts from WebVid-10M~\citep{bain2021frozen} and generated layouts with GPT-4 using the same prompts as described in our main paper. We fine-tuned the LLaMA2 models for 1000 steps with a learning rate of 2e-4 and a cosine scheduler. In addition to the visual quality (FID/FVD) and video-text alignment (CLIPSIM) scores on MSR-VTT and movement direction accuracy score on ActionBench-Directions, we also report the number of samples whose output layouts can be successfully parsed (\eg, \#Samples).
We consider the output valid if it contains 9 frames, and each frame includes at least one object with a bounding box layout. Videos are generated with $\alpha=0.1$ for MSR-VTT and $\alpha=0.2$ for ActionBench-Directions.

\subsection{Entity Grounding Embeddings: Image v.s. Text}
As discussed in \cref{sec:grounded_video_generation} in our main paper,
we compare using different embeddings for entity grounding on 1000 randomly sampled MSR-VTT test prompts.
As shown in \Cref{table:ablation_grounding_token},
CLIP image embedding is more effective than CLIP text embedding,
and using the CLIP image-text joint embedding yields the best results.
Thus, we propose to use the image+text embeddings for the default configuration.

\begin{table}[h]
    \centering
    \resizebox{0.75\linewidth}{!}{
    \begin{tabular}{l ccc c}
        \toprule
        \multirow{2}{*}{\parbox{3.5cm}{Entity Grounding}} & \multicolumn{3}{c}{MSR-VTT} & \corefdata{} \\
        \cmidrule(lr){2-4} \cmidrule(lr){5-5} 
        & FVD ($\downarrow$) & FID ($\downarrow$) & CLIPSIM ($\uparrow$) & Consistency (\%) \\
        \midrule
        Image Emb. & {737} & {18.38} & 0.2834 & 42.6  \\
        Text Emb. & {875} & {23.18} & 0.2534 & 36.9 \\
        \rowcolor{lightblue}
        Image+Text Emb. (default) & \textbf{606} & \textbf{14.60} & \textbf{0.2842} & \textbf{42.8} \\
        \bottomrule
    \end{tabular}
    }
    \caption{Ablation of entity grounding embeddings of our \videomodule{} on MSR-VTT and Coref-SV.
    }
    \label{table:ablation_grounding_token}
\vspace{-3pt}
\end{table}
\normalsize

\subsection{Layout Control: Bounding Box v.s. Center Point}
In \Cref{table:ablation_spatial_control},
we compare different layout representation formats on 1000 randomly sampled MSR-VTT test prompts.
We use image embedding for entity grounding and $\alpha=0.2$ for layout control.
Compared with no layout (`w/o Layout input') or center point-based layouts (without object shape, size, or aspect ratio),
the bounding box based layout guidance gives better visual quality (FVD/FID) and text-video alignment (CLIPSIM).

\begin{table}[h]
\begin{center}
    \resizebox{0.58\linewidth}{!}{
    \begin{tabular}{lccc}
        \toprule
        Layout representation & FVD ($\downarrow$) & FID ($\downarrow$) & CLIPSIM ($\uparrow$) \\
        \midrule
        w/o Layout input  & {639} & {15.28} & 0.2777  \\
        Center point  & {816} & {18.65} & 0.2707  \\
        \rowcolor{lightblue}
        Bounding box (default)  & \textbf{606} & \textbf{14.60} & \textbf{0.2842}  \\
        \bottomrule
    \end{tabular}
    }
\end{center}
\caption{
Ablation of layout representation of our \method{} on MSR-VTT.
}
\label{table:ablation_spatial_control}
\end{table}

\subsection{Trainable Layers: Gated Self-Attention Only v.s. Entire Guided 2D Attention}

\paragraph{Training settings.} In our \videomodule{}, only the MLP layers for grounding tokens and the gated self-attention layers are trained, accounting for 13\% of the total parameters. We conducted an ablation study to unfreeze more parameters from the \modelscope{} backbone. Specifically, all layers in the guided 2D attention module, such as self-attention, gated self-attention, and cross-attention, are made trainable, which constitutes 27\% of the total parameters. Starting from the model trained in our main paper, we further train this variant with more trainable parameters for 50k steps, keeping all other training hyper-parameters constant.

\begin{table}[h]
\begin{center}
    \resizebox{0.9\linewidth}{!}{
    \begin{tabular}{lccc}
        \toprule
        Trainable Layers & FVD ($\downarrow$) & FID ($\downarrow$) & CLIPSIM ($\uparrow$) \\
        \midrule
        Entire Guided 2D Attn (GatedSA + SelfAttn + CrossAttn)  & {1626} & {70.47} & 0.2231  \\
        \rowcolor{lightblue}
         GatedSA only (default)  & \textbf{550} & \textbf{12.22} & \textbf{0.2860}  \\
        \bottomrule
    \end{tabular}
    }
\end{center}
\caption{
Comparison of only updating parameters in gated self-attention (GatedSA) between updating all parameters in guided 2D attention (Entire Guided 2D Attn) of our \method{} on MSR-VTT. 
}
\label{table:ablation_trainable_parameters}
\end{table}

\paragraph{Result analysis.} As shown in \Cref{table:ablation_trainable_parameters}, unfreezing all parameters in the guided 2D attention module results in significantly worse performance compared to training only the additional gated self-attention layers. This decline in performance occurs because the temporal attention and temporal convolution layers, which are inactive during our image-data training, cause misalignment in the spatial and temporal layers during inference when the temporal layers are activated.

\section{Additional Experiments}
\label{appendix:additional_experiments}

\subsection{Comparison on UCF-101}

In addition to the MSR-VTT dataset for open-domain single-scene video generation, we also conduct experiments for the UCF-101 dataset~\citep{soomro2012ucf101}. Since ModelScopeT2V does not report their evaluation results on UCF-101 dataset in their technical report, we use their publicly released checkpoint and run evaluation on the same 2048 randomly sampled test prompts as used in our VideoDirectorGPT evaluation. Compared with ModelScopeT2V, we achieve a significant improvement in FVD (ModelScopeT2V 1093 vs. Ours 748) and competitive performance in the Inception Score (ModelScopeT2V 19.49 vs. Ours 19.42). In addition, we notice that ModelScopeT2V is not good at video generation on this UCF-101 dataset, which is also observed in other recent works~\citep{he2023animate}. We could expect our VideoDirectorGPT to perform better with stronger T2V backbones.

\begin{table}[h]
\centering
\resizebox{0.5\linewidth}{!}{
\begin{tabular}{l cc}
\toprule
\multirow{2}{*}{Method}  & \multicolumn{2}{c}{UCF-101} \\
\cmidrule(lr){2-3} 
& FVD ($\downarrow$) & IS ($\uparrow$) \\
\midrule
\multicolumn{2}{l}{\textcolor{gray}{Different arch / Training data}} \\
LVDM  &	917 & ${-}$  \\
CogVideo (Chinese)  & 751 & 23.55\\
CogVideo (English) & 701 & 25.27\\
MagicVideo & 699 & ${-}$\\
VideoLDM   & 550 & 33.45	\\
Make-A-Video  & 367  & 33.00\\
\midrule
\multicolumn{3}{l}{\textcolor{gray}{Same video backbone \& Test prompts}} \\
\modelscope{}$^\dagger$  & 1093 & \textbf{19.49} \\
\rowcolor{lightblue}
\method{}  & \textbf{748} & 19.42  \\
\bottomrule
\end{tabular}
}
\caption{ 
Evaluation results on UCF-101.
\modelscope{}$^\dagger$:
Our replication with 2048 randomly sampled test prompts.
}
\label{table:ucf101}
\end{table}

\section{Additional Visualization and Qualitative Examples}
\label{appendix:visualizations}

\subsection{Visualization of Different Layout-Guided Denoising Steps.}\label{appendix:denoising_steps}
\paragraph{Balancing layout control strength with visual quality.}
During video generation,
we use two-stage denoising in \videomodule{} following \citet{li2023gligen},
where we first use layout-guidance with Guided 2D attention for $\alpha*N$ steps,
and use the denoising steps without Guided 2D attention for the remaining $(1-\alpha)*N$ steps, where $N$ is the total number of denoising steps, and $\alpha \in [0,1]$ is the ratio of layout-guidance denoising steps. In our additional analysis (see main paper Table 4), we find that a high $\alpha$ generally increases layout control but could lead to lower visual quality, which is also consistent with the finding in \citet{li2023gligen}.
By default, we use $\alpha=0.1$ and $N=50$ denoising steps.
We also explore using the LLM to determine the $\alpha$ value within the range $[0, 0.3]$ during the \videoplan{} generation
(see main paper \cref{sec:additional_analysis} for details).

\begin{figure}[t]
\begin{center}
    \includegraphics[width=.88\linewidth]{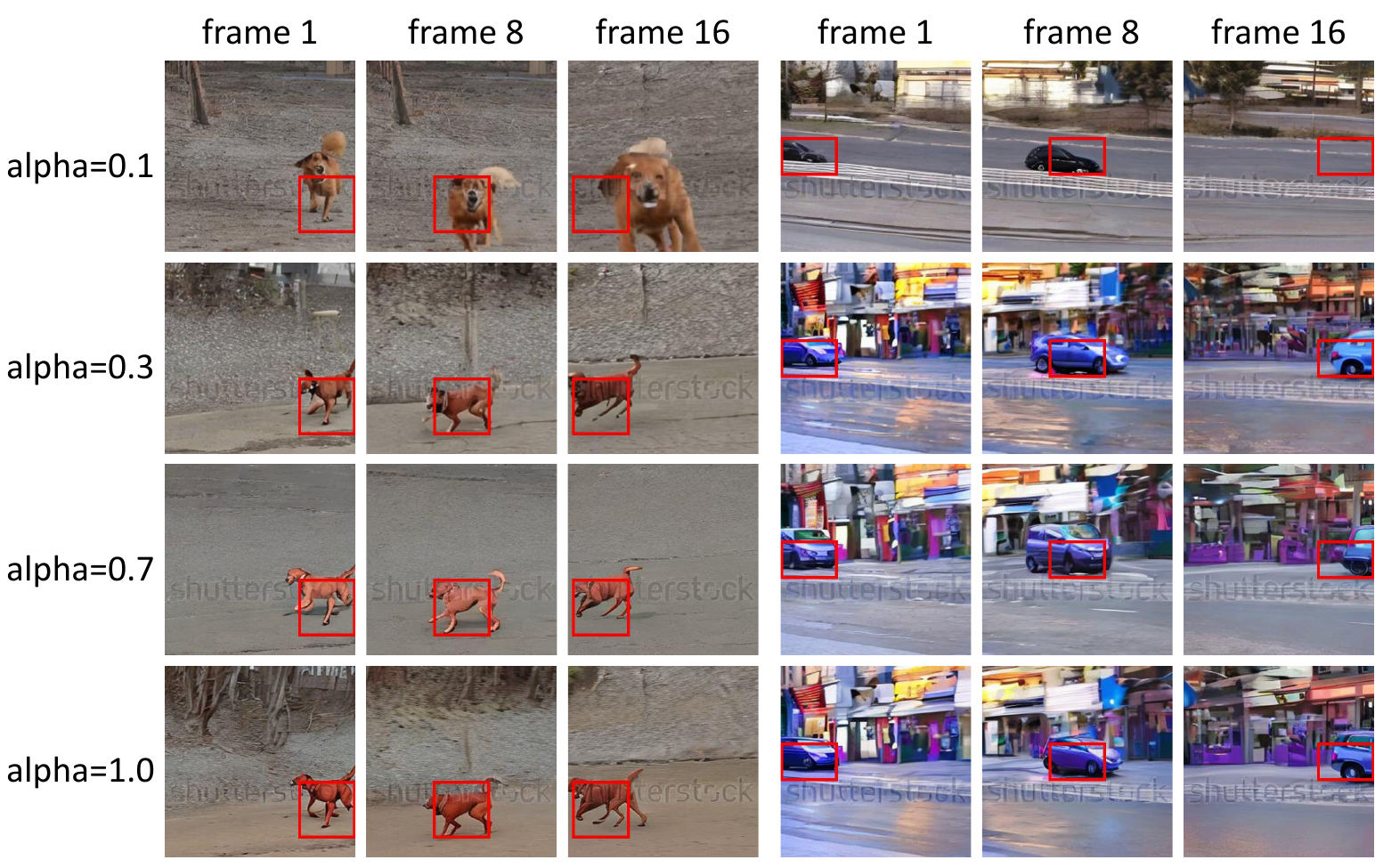}
    \vspace{-2mm}
    \caption{
       Visualization of objects with different strength of layout control (different $\alpha$ values). The prompts are ``A dog moving from right to left'' and ``A car moving from left to right'' respectively. Increasing the $\alpha$ value can make the objects better following bounding box layouts. An $\alpha$ value between 0.1 to 0.3 usually gives videos with best trade-off between visual quality and layout control.
    }
\label{fig:visualization_alpha_moveable}
\end{center}
\end{figure}

\paragraph{Visualization and Analysis.} Here, we provide visualizations of objects with prompts that involve explicit layout movements to illustrate the effect of changing the $\alpha$ hyper-parameter. As can be seen in \cref{fig:visualization_alpha_moveable}, 
a small $\alpha$ value (\eg, 0.1) can provide \textbf{guidance} for object movement, enabling the \videomodule{} to generate videos that do not strictly adhere to bounding box layouts. This approach fosters better creativity and diversity in the generated videos, and enhances visual quality as well as robustness to LLM-produced layouts.
On the other hand, an $\alpha$ value of 0.3 or higher is sufficient to \textbf{control} objects to follow bounding box trajectories. 
In conclusion, for prompts requiring explicit movement control, a larger $\alpha$ value typically results in better performance (as demonstrated by the results from ActionBench-Directions in Table 4 of our main paper). Meanwhile, for prompts that do not require large movements, a smaller 
$\alpha$ value generally leads to improved visual quality (as indicated by the results from MSR-VTT in \Cref{tab:ablation_alpha} in the main paper).

\subsection{Additional Qualitative Examples}\label{appendix:additional_qualitative_examples}

\vspace{5pt}

\begin{figure}[t]
\begin{center}
    \includegraphics[width=0.8\linewidth]{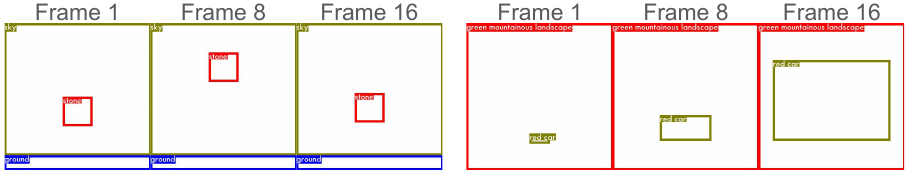}
\caption{Generated layouts from prompts that require physical understanding: ``A stone thrown into the sky'' (left) and ``A car is approaching from a distance'' (right).}
\label{fig:r1w4}
\end{center}
\end{figure}

\begin{figure}[t]
\begin{center}
    \includegraphics[width=0.99\linewidth]{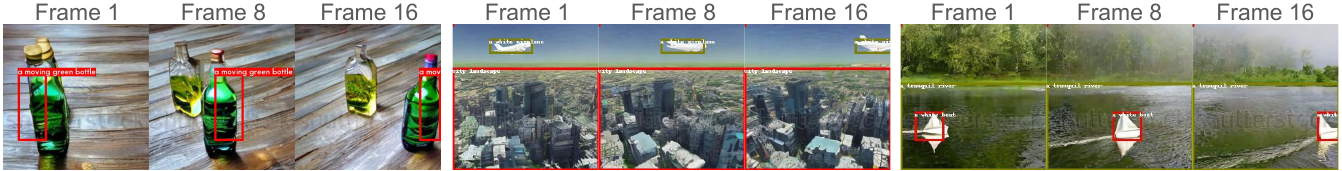}
\caption{Generated videos from movement prompts with a static object (bottle) and objects that can naturally move (airplan/boat). We use prompts ``A \{bottle/airplane/boat\} moving from left to right.''}
\label{fig:r2w2}
\end{center}
\end{figure}
\vspace{-10pt}

\begin{figure}[h!]
\begin{center}
    \includegraphics[width=0.99\linewidth]{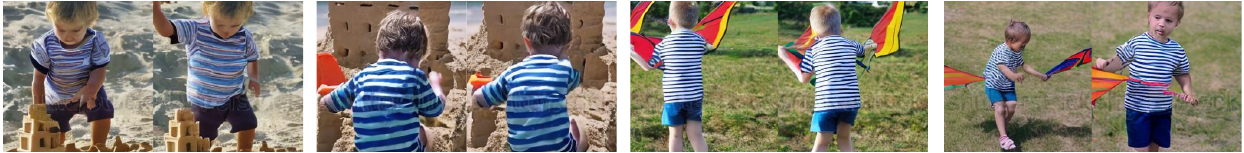}
\caption{Multi-scene video generation example from prompt: ``A boy playing in sand and flying kites.''}
\label{fig:multi_scene_boy}
\end{center}
\end{figure}

\paragraph{Prompts requiring understanding of physical world.}
{\cref{fig:r1w4}} shows that our GPT-4 based video planner
can generate object
movements requiring an understanding of the physical world such as gravity and perspective.

\paragraph{Object movement prompts with different types of objects.}
In \cref{fig:r2w2}, we show additional examples for single-scene video generation with prompts involve movements. 
For static objects (\eg{}, a bottle), movement is often depicted via the camera.
For objects that can naturally move (\eg{}, an airplane and a boat), the videos show the movements of objects.

\begin{figure}[h]
\begin{center}
    \includegraphics[width=0.99\linewidth]{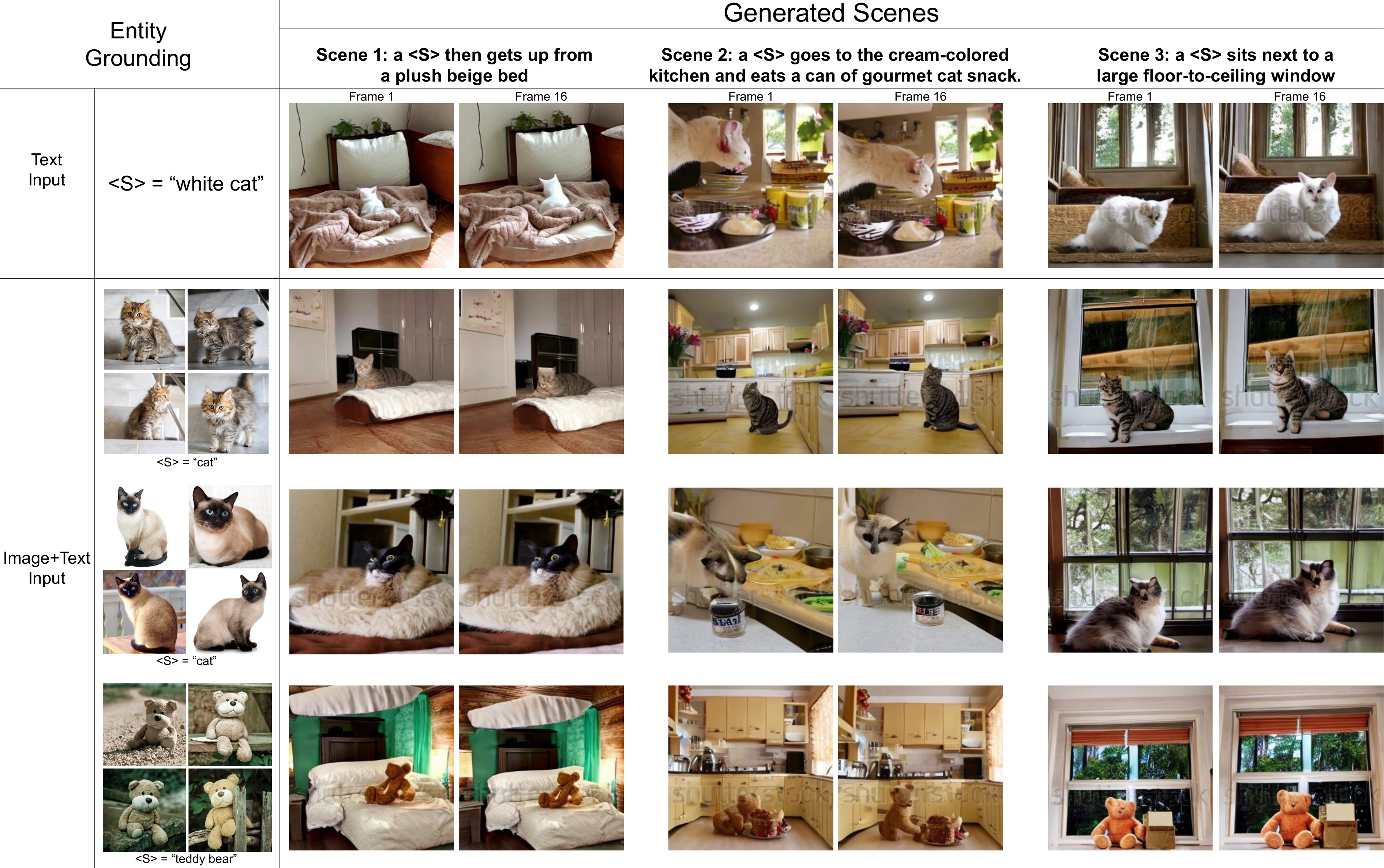}
    \vspace{-4mm}
    \caption{
    Video generation examples with \textbf{text-only and image+text inputs}.
    Users can flexibly provide either text-only or image+text descriptions to place custom entities when generating videos with \method{}.
    For both text and image+text based entity grounding examples,
    the identities of the provided entities are well preserved across multiple scenes. 
    }
\label{fig:image_input}
\end{center}
\end{figure}

\paragraph{Generating videos with custom image exemplars.}
In \cref{fig:image_input}, we demonstrate that our Layout2Vid can flexibly take either text-only or
image+text descriptions as input to generate multi-scene videos with good entity consistency.

\paragraph{Multi-scene video generation with human appearance.}
In \cref{fig:multi_scene_boy}, we show an additional example of multi-scene video generation with a human appearance. Our \method{} can well preserve the boy's appearances across scenes.

\begin{figure}[h]
\begin{center}
    \includegraphics[width=.85\linewidth]{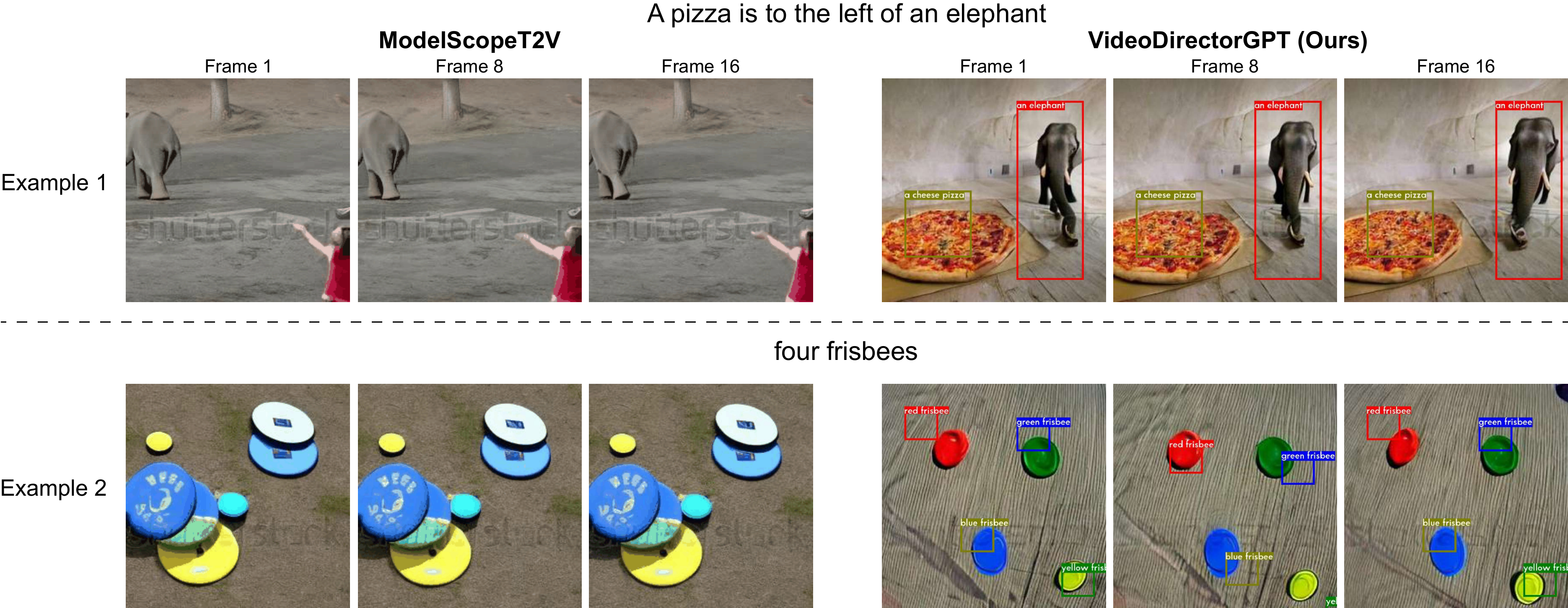}
    \caption{
        Video generation examples on \textbf{VPEval Skill-based prompts} for spatial and count skills.
        Our \videoplan{}, with object layouts overlaid, successfully guides the \videomodule{} module to place objects in the correct spatial relations and to depict the correct number of objects,
        whereas \modelscope{} fails to generate a `pizza' in the first example and overproduces the number of frisbees in the second example. 
    }
\label{fig:vpeval_outputs}
\vspace{-2mm}
\end{center}
\end{figure}

\paragraph{VPEval Skill-based.}
\cref{fig:vpeval_outputs} shows that LLM-generated video plan successfully guides the Layout2Vid module to accurately place objects in the correct
spatial relations and to generate the correct number of objects.
In contrast, ModelScopeT2V fails to generate
a ‘pizza’ in the first example and overproduces the number of frisbees in the second example.

\begin{figure}[h]
\begin{center}
    \includegraphics[width=.85\linewidth]{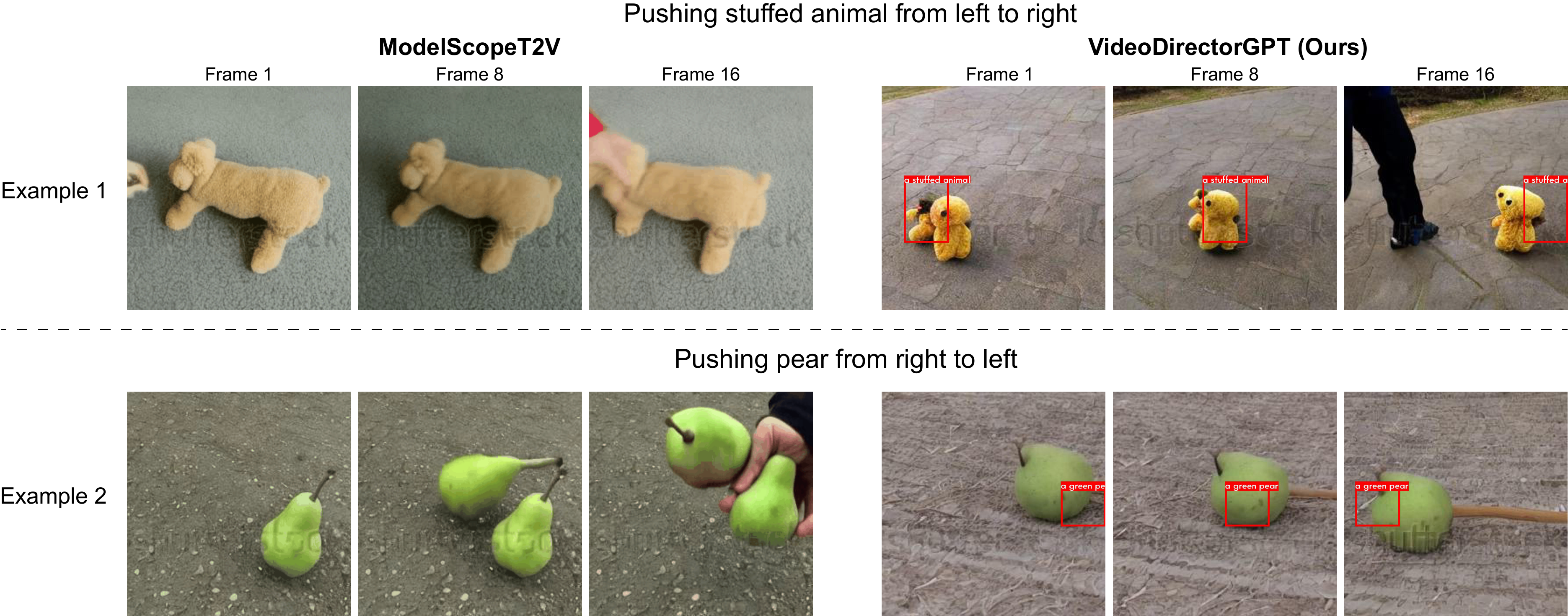}
    \vspace{-2mm}
    \caption{
        Video generation examples on \textbf{ActionBench-Direction prompts}.
        Our \videoplan{}'s object layouts (overlaid) can guide the \videomodule{} module to place and move the `stuffed animal' and `pear' in their correct respective directions,
        whereas the objects in the \modelscope{} videos stay in the same location or move in random directions. 
    }
\label{fig:actionbench_outputs}
\vspace{-2mm}
\end{center}
\end{figure}

\paragraph{ActionBench-direction.}
\cref{fig:actionbench_outputs} shows that
our LLM-generated video plan can guide the Layout2Vid module to place the ‘stuffed animal’ and the ‘pear’
in their correct starting positions and then move them toward the correct end positions, whereas the objects
in the ModelScopeT2V videos stay in the same location or move in random directions.

\begin{figure}[h]
\begin{center}
    \includegraphics[width=.88\linewidth]{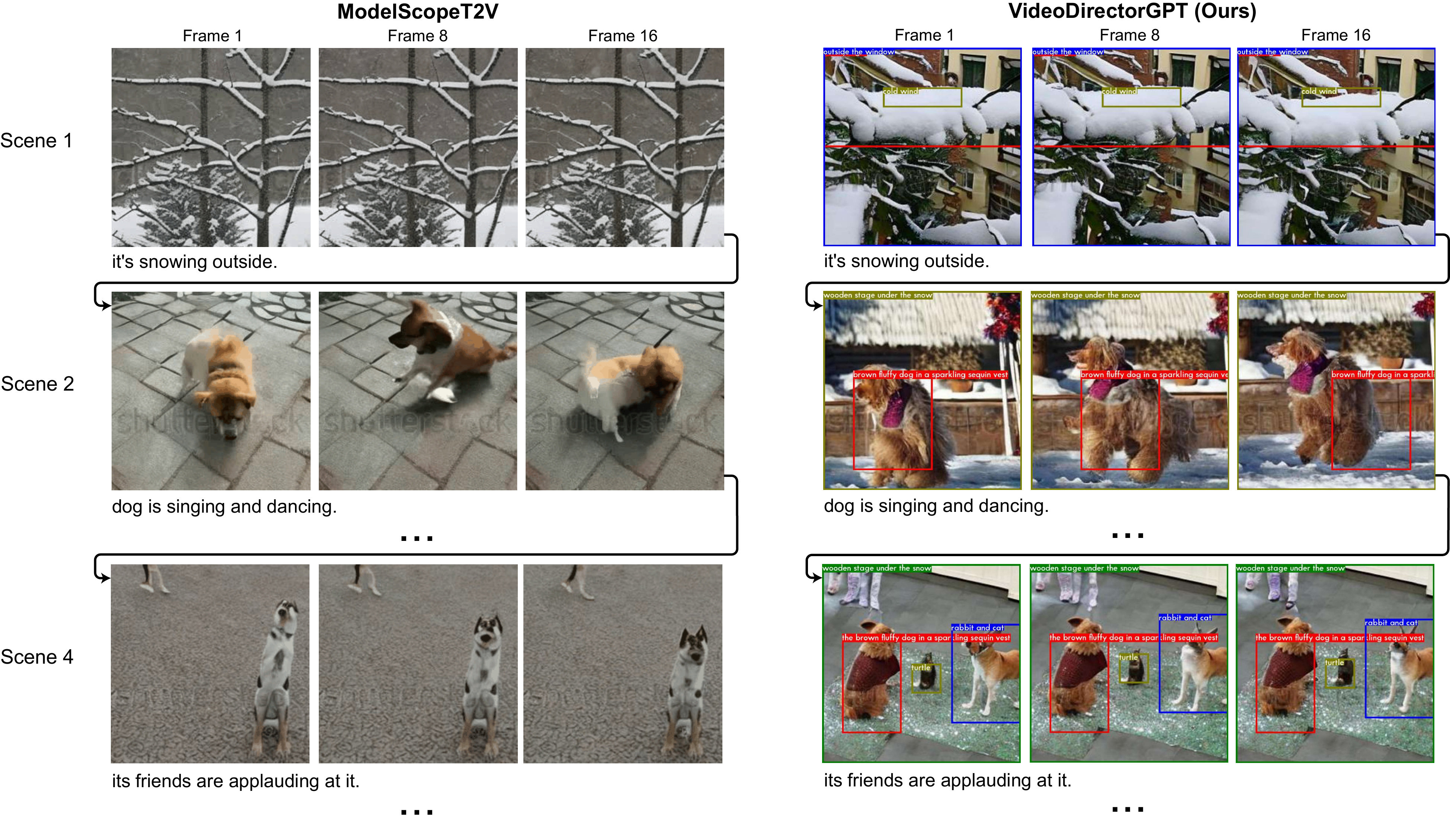}
    \vspace{-2mm}
    \caption{
        Video generation examples on \textbf{\corefdata{} prompts}.
        Our \videoplan{}'s object layouts (overlaid) can guide the \videomodule{} module to generate the same brown dog and maintain snow across scenes consistently,
        whereas \modelscope{} generates different dogs in different scenes and loses the snow after the first scene. 
    }
\label{fig:pororo_outputs}
\vspace{-2mm}
\end{center}
\end{figure}

\paragraph{\corefdata{}.}
\cref{fig:pororo_outputs} shows that 
our \videoplan{} guide the \videomodule{} module to generate the same dog and maintain snow across scenes consistently,
whereas \modelscope{} generates different dogs in different scenes and loses the snow after the first scene.

\begin{figure}[h]
\begin{center}
    \includegraphics[width=.88\linewidth]{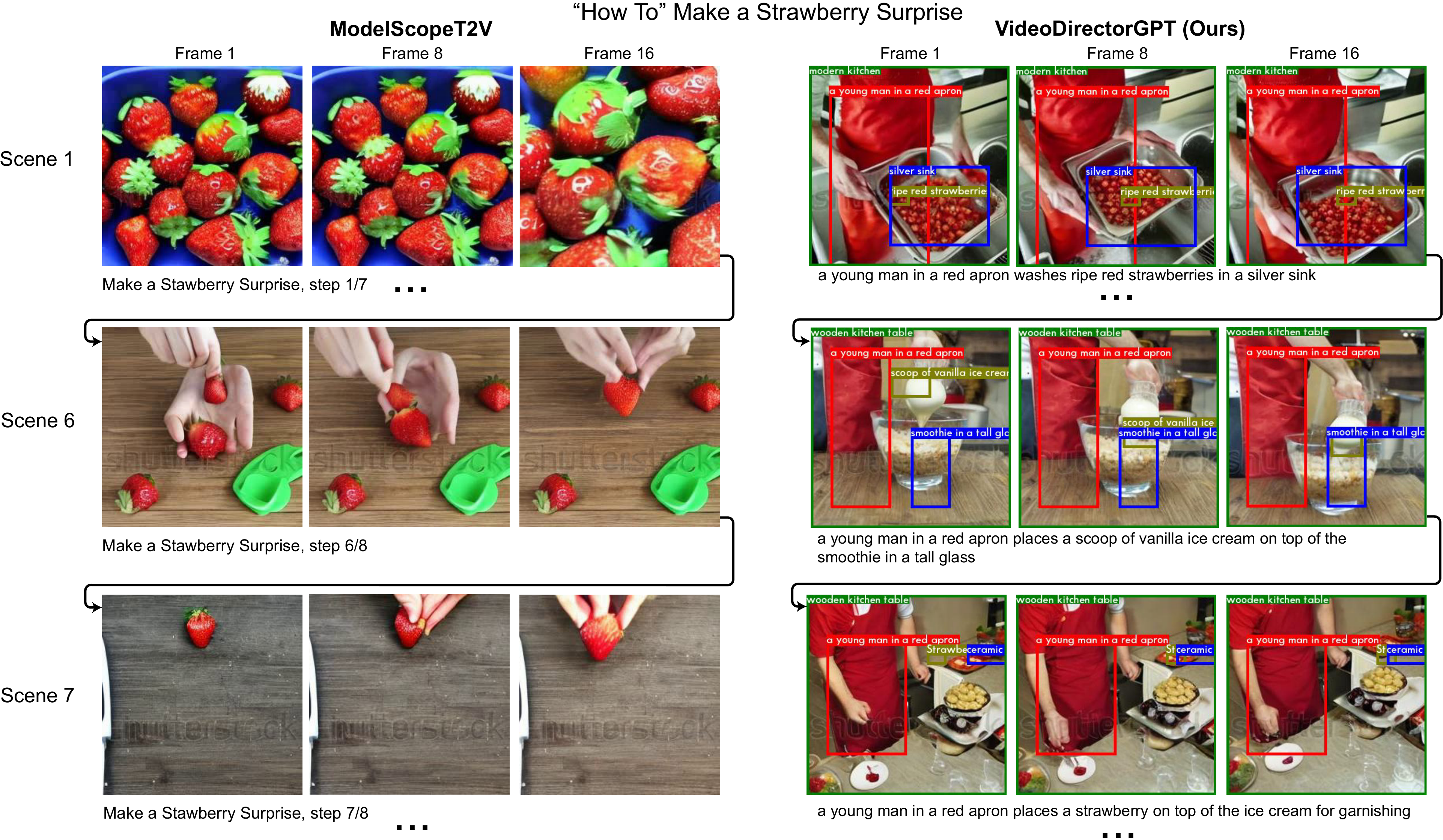}
    \vspace{-2mm}
    \caption{
        Video generation examples \textbf{HiREST prompts}. Our \method{} generates a detailed \videoplan{} that properly expands the original text prompt, ensures accurate object bounding box locations (overlaid), and maintains consistency of the person across the scenes. 
        \modelscope{} only generates strawberries, without strawberry surprise. 
    }
\label{fig:hirest_outputs_2}
\vspace{-2mm}
\end{center}
\end{figure}

\paragraph{HiREST.}
\cref{fig:hirest_outputs_2} shows that
our LLM can generate step-by-step \videoplan{} from a single prompt, and our \videomodule{} can generate consistent videos following the plan.
Our \method{} breaks down the process and generates a complete video showing how to make a strawberry surprise (a type of dessert consisting of vanilla ice cream and strawberries). \modelscope{} repeatedly generates strawberries.

\section{Limitations}
\label{sec:limitations}

While our framework can be beneficial for numerous applications (\eg{}, user-controlled/human-in-the-loop video generation/manipulation and data augmentation), akin to other video generation frameworks, it can also be utilized for potentially harmful purposes (\eg{}, creating false information or misleading videos), and thus should be used with caution in real-world applications (with human supervision, \eg{}, as described in \Cref{appendix:human_in_the_loop} --- human-in-the-loop video plan editing).
Also, generating a \videoplan{} using the strongest LLM APIs can be costly,
similar to other recent LLM-based frameworks.
We hope that advances in quantization/distillation and open-source models will continue to lower the inference cost of LLMs.
Lastly, our video generation module (\videomodule{}) is based on the pretrained weights of a specific T2V generation backbone. Therefore, we face similar limitations to their model, including deviations related to the distribution of training datasets, imperfect generation quality, and only understanding the English corpus.
\vspace{20pt}

\end{document}